\documentclass{article}
\usepackage[bookmarks=false]{hyperref}
\usepackage[numbers]{natbib}
\usepackage[final]{nips_2018}
\usepackage{algorithm,algorithmic,graphicx,graphicx,amsmath,amsfonts,amssymb,multirow,xcolor,colortbl,dsfont,amsthm}
\usepackage{wrapfig}
\usepackage{times}
\makeatletter
\renewenvironment{proof}[1][\proofname]{\par
 \pushQED{\qed}%
 \normalfont \topsep6\p@\@plus6\p@\relax
 \trivlist
 \item[\hskip\labelsep
 \bfseries
 #1\@addpunct{.}]\ignorespaces
}{%
 \popQED\endtrivlist\@endpefalse
}
\makeatother
\def\eop{$\rule{1.3ex}{1.3ex}$}
\renewcommand\qedsymbol\eop 
\newtheorem{definition}{Definition}
\newtheorem{theorem}{Theorem}
\newtheorem{proposition}{Proposition}
\usepackage[utf8]{inputenc}
\usepackage[T1]{fontenc} 
\author{
 Michele Donini \\
 CSML \\
 Istituto Italiano di Tecnologia \\
 Genoa, Italy \\
 \texttt{michele.donini@iit.it} 
 \And
 Luca Oneto \\
 DIBRIS \\
 University of Genova \\
 Genoa, Italy \\
 \texttt{luca.oneto@unige.it} 
 \And
 Shai Ben-David \\
 School of Computer Science \\
 University of Waterloo \\
 Waterloo, Ontario, Canada \\
 \texttt{shai@cs.uwaterloo.ca} 
 \And
 John Shawe-Taylor \\
 Department of Computer Science \\
 University College London \\
 London, UK \\
 \texttt{j.shawe-taylor@ucl.ac.uk} 
 \And
 Massimiliano Pontil \\
 CSML \\
 Istituto Italiano di Tecnologia \\
 Genoa, Italy \\
 \texttt{massimiliano.pontil@iit.it} 
}
\title{Empirical Risk Minimization \\ Under Fairness Constraints}
\begin{document}
\maketitle
\begin{abstract}
We address the problem of algorithmic fairness: ensuring that sensitive variables do not unfairly influence the outcome of a classifier.
We present an approach based on empirical risk minimization, which incorporates a fairness constraint into the learning problem.
It encourages the conditional risk of the learned classifier to be approximately constant with respect to the sensitive variable.
We derive both risk and fairness bounds that support the statistical consistency of our approach.
We specify our approach to kernel methods and observe that the fairness requirement implies an orthogonality constraint which can be easily added to these methods.
We further observe that for linear models the constraint translates into a simple data preprocessing step.
Experiments indicate that the method is empirically effective and performs favorably against state-of-the-art approaches.
\end{abstract}
\section{Introduction}
\label{sec:intro}
In recent years there has been a lot of interest on algorithmic fairness in machine learning~see, e.g.,~\cite{dwork2018decoupled,hardt2016equality,zafar2017fairness,zemel2013learning,kilbertus2017avoiding,kusner2017counterfactual,calmon2017optimized,joseph2016fairness,chierichetti2017fair,jabbari2016fair,yao2017beyond,lum2016statistical,zliobaite2015relation} and references therein.
The central question is how to enhance supervised learning algorithms with fairness requirements, namely ensuring that sensitive information (e.g.~knowledge about the ethnic group of an individual) 
does not `unfairly' influence the outcome of a learning algorithm.
For example if the learning problem is to decide whether a person should be offered a loan based on her previous credit card scores, we would like to build a model which does not unfair use additional sensitive information such as race or sex.
Several notions of fairness and associated learning methods have been introduced in machine learning in the past few years, including Demographic Parity~\cite{calders2009building}, Equal Odds and Equal Opportunities~\cite{hardt2016equality}, Disparate Treatment, Impact, and mistreatment~\cite{zafar2017fairness}.
The underlying idea behind such notions is to balance decisions of a classifier among the different sensitive groups and label sets.

In this paper, we build upon the notion Equal Opportunity (EO) which defines fairness as the requirement that the true positive rate of the classifier is the same across the sensitive groups.
In Section~\ref{sec:luca:th:Fairness} we introduce a generalization of this notion of fairness which constrains the conditional risk of a classifier, associated to positive labeled samples of a group, to be approximately constant with respect to group membership.
The risk is measure according to a prescribed loss function and approximation parameter $\epsilon$.
When the loss is the misclassification error and $\epsilon = 0$ we recover the notion EO above.
We study the problem of minimizing the expected risk within a prescribed class of functions subject to the fairness constraint.
As a natural estimator associated with this problem, we consider a modified version of Empirical Risk Minimization (ERM) which we call Fair ERM (FERM).
We derive both risk and fairness bounds, which support that FERM is statistically consistent, in a certain sense which we explain in the paper in Section~\ref{sec:luca:th:FERM}.
Since the FERM approach is impractical due to the non-convex nature of the constraint, we propose, still in Section~\ref{sec:luca:th:FERM}, a surrogate convex FERM problem which relates, under a natural condition, to the original goal of minimizing the misclassification error subject to a relaxed EO constraint.
We further observe that our condition can be empirically verified to judge the quality of the approximation in practice.
As a concrete example of the framework, in Section~\ref{sec:luca:th:FK} we describe how kernel methods such as support vector machines (SVMs) can be enhanced to satisfy the fairness constraint.
We observe that a particular instance of the fairness constraint for $\epsilon=0$ reduces to an orthogonality constraint.
Moreover, in the linear case, the constraint translates into a preprocessing step that implicitly imposes the fairness requirement on the data, making fair any linear model learned with them.
We report numerical experiments using both linear and nonlinear kernels, which indicate that our method improves on the state-of-the-art in four out of five datasets and is competitive on the fifth dataset\footnote{
Additional technical steps and experiments are presented in the supplementary materials.}.


In summary the contributions of this paper are twofold. First we outline a general framework for empirical risk minimization under fairness constraints. The framework can be used as a starting point to develop specific algorithms for learning under fairness constraints. As a second contribution, we shown how a linear fairness constraint arises naturally in the framework and allows us to develop a novel convex learning method that is supported by consistency properties both in terms of EO and risk of the selected model, performing favorably against state-of-the-art alternatives on a series of benchmark datasets.

\noindent {\bf Previous Work.} Work on algorithmic fairness
can be divided in three families. Methods in the first family modify a pretrained classifier in order to increase its fairness properties while maintaining as much as possible the classification performance:~\cite{pleiss2017fairness,beutel2017data,hardt2016equality,feldman2015certifying} are examples of these methods but no consistency property nor comparison with state-of-the-art proposal are provided.
Methods in the second family enforce fairness directly during the training step:~\cite{agarwal2017reductions,agarwal2018reductions,woodworth2017learning,zafar2017fairness,menon2018cost,zafar2017parity,bechavod2018Penalizing,zafar2017fairnessARXIV,kamishima2011fairness,kearns2017preventing} are examples of this method which provide non-convex approaches to the solution of the problem or they derive consistency results just for the non-convex formulation resorting later to a convex approach which is not theoretically grounded;~\cite{Prez-Suay2017Fair,dwork2018decoupled,berk2017convex,alabi2018optimizing} are other examples of convex approaches which do not compare with other state-of-the-art solutions and do not provide consistency properties except for the~\cite{dwork2018decoupled} which, contrarily to our proposal, does not enforce a fairness constraint directly in the learning phase and the~\cite{olfat2018spectral} which proposes a computational tractable fair SVM starting from a constraint on the covariance matrices. Specifically, it leads to a non-convex constraint which is imposed iteratively with a sequence of relaxation exploiting spectral decompositions.
Finally, the third family of methods implements fairness by modifying the data representation and then employs standard machine learning methods:~\cite{adebayo2016iterative,calmon2017optimized,kamiran2009classifying,zemel2013learning,kamiran2012data,kamiran2010classification} are examples of these methods but, again, no consistency property nor comparison with state-of-the-art proposal are provided.
Our method belongs to the second family of methods, in that it directly optimizes a fairness constraint related to the notion of EO discussed above.
Furthermore, in the case of linear models, our method translates to an efficient preprocessing of the input data, with methods in the third family.
As we shall see, our approach is theoretically grounded and performs favorably against the state-of-the-art\footnote{A detailed comparison between our proposal and state-of-the-art is reported in the supplementary materials.}.
%
%
\section{Fair Empirical Risk Minimization}
\label{sec:luca:th:Fairness}
In this section, we present our approach to learning with fairness. We begin by introducing our notation. We let $\mathcal{D} = \left\{ (\boldsymbol{x}_1,s_1,y_1),\dots, (\boldsymbol{x}_n,s_n,y_n) \right\}$ be a sequence of $n$ samples drawn independently from an unknown probability distribution $\mu$ over $\mathcal{X} \times \mathcal{S} \times \mathcal{Y}$, where $\mathcal{Y} = \{ -1, +1 \}$ is the set of binary output labels, $\mathcal{S} = \{a,b\}$ represents group membership among two groups\footnote{The extension to multiple groups (e.g.~ethnic group) is briefly discussed in the supplementary material.} (e.g.~`female' or `male'), and $\mathcal{X}$ is the input space.
We note that the $\boldsymbol{x} \in \mathcal{X}$ may further contain or not the sensitive feature $s \in \mathcal{S}$ in it.
We also denote by $\mathcal{D}^{+,g} {=}\{(\boldsymbol{x}_i,s_i,y_i) : y_i {=} 1,s_i {=} g \}$ for $g \in \{ a,b \}$ and $n^{+,g} = |\mathcal{D}^{+,g}|$.
Let us consider a function (or model) $f: \mathcal{X} \rightarrow \mathbb{R}$ chosen from a set $\mathcal{F}$ of possible models.
The error (risk) of $f$ in approximating $\mu$ is measured by a prescribed loss function $\ell:\mathbb{R} \times \mathcal{Y} \rightarrow \mathbb{R}$.
The risk of $f$ is defined as ${L}(f) = \mathbb{E} \left[ \ell(f(\boldsymbol{x}),y) \right]$.
When necessary we will indicate with a subscript the particular loss function used, i.e.~${L}_p(f) = \mathbb{E} \left[ \ell_p(f(\boldsymbol{x}),y) \right]$.

The purpose of a learning procedure is to find a model that minimizes the risk.
Since the probability measure $\mu$ is usually unknown, the risk cannot be computed, however we can compute the empirical risk $\hat{L}(f) = \hat{\mathbb{E}} [\ell(f(\boldsymbol{x}),y)]$, where $\hat{\mathbb{E}}$ denotes the empirical expectation.
A natural learning strategy, called Empirical Risk Minimization (ERM), is then to minimize the empirical risk within a prescribed set of functions.
\subsection{Fairness Definitions}
\label{sec:luca:th:Definitions}
In the literature there are different definitions of fairness of a model or learning algorithm~\cite{hardt2016equality,dwork2018decoupled,zafar2017fairness,zafar2017fairness}, but there is not yet a consensus on which definition is most appropriate.
In this paper, we introduce a general notion of fairness which encompasses some previously used notions and it allows to introduce new ones by specifying the loss function used below.
\begin{definition}
\label{def:fairness}
Let ${L}^{+,g}(f) {=} \mathbb{E} [ \ell(f(\boldsymbol{x}),y) | y {=} 1, s {=} g ]$ be the risk of the positive labeled samples in the $g$-th group, and let $\epsilon \in [0,1]$.
We say that a function $f$ is $\epsilon$-fair if ~$| {L}^{+,a}(f) - {L}^{+,b}(f)| \leq \epsilon$.
\end{definition}
This definition says that a model is fair if it commits approximately the same error on the positive class independently of the group membership.
That is, the conditional risk $L^{+,g}$ is approximately constant across the two groups.
Note that if $\epsilon = 0$ and we use the hard loss function, $\ell_h(f(\boldsymbol{x}),y) = \mathds{1}_{\{y f(\boldsymbol{x}) \leq 0\}}$, then Definition~\ref{def:fairness} is equivalent to definition of EO proposed by~\cite{hardt2016equality}, namely
\begin{align}
\mathbb{P}\left\{ f(\boldsymbol{x}) > 0 ~|~ y = 1, s = a \right\} =
\mathbb{P}\left\{ f(\boldsymbol{x}) > 0 ~|~ y = 1, s = b \right\}.
\label{eq:DEO}
\end{align}
This equation means that the true positive rate is the same across the two groups.
Furthermore, if we use the linear loss function 
$\ell_l(f(\boldsymbol{x}),y) = (1 - y f(\boldsymbol{x}))/2 $ and set $\epsilon = 0$, then Definition~\ref{def:fairness} gives
\begin{align}
\mathbb{E}[f(\boldsymbol{x}) ~|~ y = 1, s = a] = \mathbb{E}[f(\boldsymbol{x}) ~|~ y = 1, s = b ].
\label{eq:lollo}
\end{align}
By reformulating this expression we obtain a notion of fairness that has been proposed by~\cite{dwork2018decoupled}
\begin{align}
\sum_{g \in \{a,b\}} \big| \mathbb{E}[f(\boldsymbol{x}) ~|~ y = 1, s = g] - \mathbb{E}[f(\boldsymbol{x}) ~|~ y = 1] \big| = 0.
\nonumber
\end{align}
Yet another implication of Eq.~\eqref{eq:lollo} is that the output of the model is uncorrelated with respect to the group membership conditioned on the label being positive, that is, for every $g {\in} \{ a, b \}$, we have
\begin{align}
\mathbb{E}\big[ f(\boldsymbol{x}) \mathds{1}_{\{s{=}g\}}~|~y=1 \big] = \mathbb{E} \big[f(\boldsymbol{x})|y=1\big] \mathbb{E} \big[\mathds{1}_{\{s=g\}}~|~y=1\big].
\nonumber
\end{align}
Finally, we observe that our approach naturally generalizes to other fairness measures, e.g.~equal odds~\cite{hardt2016equality}, which could be subject of future work. 
Specifically, we would require in Definition~\ref{def:fairness} that $| {L}^{y,a}(f) - {L}^{y,b}(f)| \leq \epsilon$ for both $y \in \{-1,1\}$.
\subsection{Fair Empirical Risk Minimization}
\label{sec:luca:th:FERM}
In this paper, we aim at minimizing the risk subject to a fairness constraint.
Specifically, we consider the problem
\begin{align}
 \min\Big\{L(f) : f {\in} \mathcal{F} ,~
\big| {L}^{+,a}(f) - {L}^{+,b}(f)\big| \leq \epsilon\Big\}
\label{eq:alg:deterministic},
\end{align}
where $\epsilon \in [0,1]$ is the amount of unfairness that we are willing to bear.
Since the measure $\mu$ is unknown we replace the deterministic quantities with their empirical counterparts.
That is, we replace Problem~\ref{eq:alg:deterministic} with
\begin{align}
\min\Big\{\hat{L}(f) : f {\in} \mathcal{F} ,~
\big| \hat{L}^{+,a}(f) - \hat{L}^{+,b}(f)\big| \leq \hat{\epsilon}\Big\}
\label{eq:alg:empirical},
\end{align}
where $\hat{\epsilon} \in [0,1]$.
We will refer to~Problem~\ref{eq:alg:empirical} as FERM.

We denote by $f^*$ a solution of Problem~\ref{eq:alg:deterministic}, and by $\hat{f}$ a solution of Problem~\ref{eq:alg:empirical}.
In this section we will show that these solutions are linked one to another.
In particular, if the parameter $\hat{\epsilon}$ is chosen appropriately, we will show that, in a certain sense, the estimator $\hat{f}$ is consistent.
In order to present our observations, we require that it holds with probability at least $1-\delta$ that
\begin{align}
\sup_{f \in \mathcal{F}} \big|L(f) - \hat{L}(f)\big| \leq B(\delta,n,\mathcal{F})
\label{eq:bartlett}
\end{align}
where the bound $B(\delta,n,\mathcal{F})$ goes to zero as $n$ grows to infinity if the class $\mathcal{F}$ is learnable with respect to the loss~\cite[see e.g.][and references therein]{shalev2014understanding}.
For example, if $\mathcal{F}$ is a compact subset of linear separators in a Reproducing Kernel Hilbert Space (RKHS), and the loss is Lipschitz in its first argument, then $B(\delta,n,\mathcal{F})$ can be obtained via Rademacher bounds~\cite[see e.g.][]{bartlett2002rademacher}.
In this case $B(\delta,n,\mathcal{F})$ goes to zero at least as ${\sqrt{1/n}}$ as $n$ grows to infinity, where $n = |\mathcal{D}|$.

We are ready to state the first result of this section (proof is reported in supplementary materials).
\begin{theorem}
\label{thm:mainresult1}
Let $\mathcal{F}$ be a learnable set of functions with respect to the loss function $\ell: \mathbb{R} \times {\cal Y} \rightarrow \mathbb{R}$, let $f^*$ be a solution of Problem (\ref{eq:alg:deterministic}) and let $\hat{f}$ be a solution of Problem (\ref{eq:alg:empirical}) with 
\begin{align}
\textstyle
\hat{\epsilon} = \epsilon + \sum_{g \in \{a,b\}} B(\delta,n^{+,g},\mathcal{F}).
\end{align}
With probability at least $1-6 \delta$ it holds simultaneously that
\begin{align}
\textstyle
L(\hat{f}) - L(f^*) \leq 2 B(\delta,n,\mathcal{F}) \quad
\text{and} \quad
\textstyle
\Big| L^{+,a}(\hat{f}) - L^{+,b}(\hat{f}) \Big| \leq \epsilon + 2 \sum_{g \in \{a,b\}} B(\delta,n^{+,g},\mathcal{F}).
\nonumber
\end{align}
\end{theorem}

A consequence of the first statement of Theorem~\ref{thm:mainresult1} is that as $n$ tends to infinity $L(\hat{f})$ tends to a value which is not larger than $L(f^*)$, that is, FERM is consistent with respect to the risk of the selected model.
The second statement of Theorem~\ref{thm:mainresult1}, instead, implies that as $n$ tends to infinity we have that $\hat{f}$ tends to be $\epsilon$-fair.
In other words, FERM is consistent with respect to the fairness of the selected model.

Thanks to Theorem~\ref{thm:mainresult1} we can state that $f^{*}$ is close to $\hat{f}$ both in term of its risk and its fairness.
Nevertheless, our final goal is to find an $f^*_h$ which solves the following problem
\begin{align}\label{eq:problemHard}
\min\Big\{L_h(f) : f {\in} \mathcal{F} ,~
\big| {L}^{+,a}_h(f) - {L}^{+,b}_h(f)\big| \leq \epsilon\Big\}.
\end{align}
Note that the objective function in Problem~\ref{eq:problemHard} is the misclassification error of the classifier $f$, whereas the fairness constraint is a relaxation of the EO constraint in Eq.~\eqref{eq:DEO}.
Indeed, the quantity $\big| {L}^{+,a}_h(f) - {L}^{+,b}_h(f)\big|$ is equal to
\begin{align}
\!\!\big| \mathbb{P}\left\{ f(\boldsymbol{x}) > 0 ~|~ y = 1,\! s = a \right\}\! - 
\mathbb{P}\left\{ f(\boldsymbol{x}) > 0 ~|~ y = 1,\! s = b \right\}\! \big|.
\label{def:DEOmichele}
\end{align}
We refer to this quantity as difference of EO (DEO).

Although Problem~\ref{eq:problemHard} cannot be solved, by exploiting Theorem~\ref{thm:mainresult1} we can safely search for a solution $\hat{f}_h$ of its empirical counterpart
\begin{align}\label{eq:problemHardempirical}
\min\Big\{\hat{L}_h(f) : f {\in} \mathcal{F} ,~
\big| \hat{L}^{+,a}_h(f) - \hat{L}^{+,b}_h(f)\big| \leq \hat{\epsilon}\Big\}.
\end{align}
Unfortunately~Problem~\ref{eq:problemHardempirical} is a difficult nonconvex nonsmooth problem, and for this reason it is more convenient to solve a convex relaxation.
That is, we replace the hard loss in the risk with a convex loss function $\ell_c$ (e.g.~the Hinge loss $\ell_{c} = \max\{0, \ell_l \}$) and the hard loss in the constraint with the linear loss $\ell_l$.
In this way, we look for a solution $\hat{f}_c$ of the convex FERM problem
\begin{align}\label{eq:problemSoft}
\min\Big\{\hat{L}_c(f) : f {\in} \mathcal{F} ,~
\big| \hat{L}^{+,a}_l(f) - \hat{L}^{+,b}_l(f)\big| \leq \hat{\epsilon}\Big\}.
\end{align}

The questions that arise here are whether, and how close, $\hat{f}_c$ is to $\hat{f}_h$, how much, and under which assumptions.
The following theorem sheds some lights on these issues (proof is reported in supplementary materials, Section~\ref{sec:SMproofs}).
\begin{proposition}
\label{thm:mainresult2}
If $\ell_c$ is the Hinge loss then $
\hat{L}_{h}(f) \leq \hat{L}_{c}(f)$.
Moreover, if for $f: \mathcal{X} \rightarrow \mathbb{R}$ the following condition is true
\begin{align} \label{eq:hp1}
\textstyle
\frac{1}{2} \sum_{g \in \{a,b\}} \left| \hat{\mathbb{E}} \left[ \operatorname{sign}\big(f(\boldsymbol{x})\big)- f(\boldsymbol{x}) ~\big|~ y = 1, s = g \right] \right| \leq \hat{\Delta},
\end{align}
then it also holds that 
\begin{align}
\textstyle
\big| \hat{L}^{+,a}_h(f) - \hat{L}^{+,b}_h(f) \big| \leq \big| \hat{L}^{+,a}_l(f) - \hat{L}^{+,b}_l(f)\big| + \hat{\Delta}.
\nonumber
\end{align}
\end{proposition}

The first statement of Proposition~\ref{thm:mainresult2} tells us that exploiting $\ell_c$ instead of $\ell_h$ is a good approximation if $\hat{L}_{c}(\hat{f}_c)$ is small.
The second statement of Proposition~\ref{thm:mainresult2}, instead, tells us that if the hypothesis of inequality (\ref{eq:hp1}) holds, then the linear loss based fairness is close to the EO.
Obviously the smaller $\hat{\Delta}$ is, the closer they are.
Inequality (\ref{eq:hp1}) says that the functions $\operatorname{sign}\big(f(\boldsymbol{x})\big)$ and $ f(\boldsymbol{x})$ distribute, on average, in a similar way.
This condition is quite natural and it has been exploited in previous work~\cite[see e.g.][]{maurer2004note}.
Moreover, in Section~\ref{sec:exps} we present experiments showing that $\hat{\Delta}$ is small.

The bound in Proposition~\ref{thm:mainresult2} may be tighten by using different nonlinear approximations of EO~\citep[see e.g.][]{calmon2017optimized}.
However, the linear approximation proposed in this work gives a convex problem, and as we shall see in Section 5, works well in practice.

In summary, the combination of Theorem~\ref{thm:mainresult1} and Proposition~\ref{thm:mainresult2} provides conditions under which a solution $\hat{f}_c$ of Problem~\ref{eq:alg:empirical}, which is convex, is close, {\em both in terms of classification accuracy and fairness}, to a solution $f^*_h$ of Problem~\ref{eq:problemHard}, which is our final goal.
\section{Fair Learning with Kernels}
\label{sec:luca:th:FK}
In this section, we specify the FERM framework to the case that the underlying space of models is a reproducing kernel Hilbert space (RKHS)~\cite[see e.g.][and references therein]{shawe2004kernel,smola2001}.
We let $\kappa: \mathcal{X} \times \mathcal{X} \rightarrow \mathbb{R}$ be a positive definite kernel and let $\boldsymbol{\phi}: \mathcal{X} \rightarrow \mathbb{H}$ be an induced feature mapping such that $\kappa(\boldsymbol{x},\boldsymbol{x}') = \langle \boldsymbol{\phi}(\boldsymbol{x}),\boldsymbol{\phi}(\boldsymbol{x}')\rangle$, for all $\boldsymbol{x},\boldsymbol{x}' \in \mathcal{X}$, where $\mathbb{H}$ is the Hilbert space of square summable sequences.
Functions in the RKHS can be parametrized as 
\begin{equation}
f(\boldsymbol{x}) = \langle \boldsymbol{w} , \boldsymbol{\phi}(\boldsymbol{x})\rangle,~~~\boldsymbol{x} \in \mathcal{X},
\label{eq:222}
\end{equation}
for some vector of parameters $\boldsymbol{w} \in \mathbb{H}$.
In practice a bias term (threshold) can be added to $f$ but to ease our presentation we do not include it here.

We solve Problem~\eqref{eq:problemSoft} with $\mathcal{F}$ a ball in the RKHS and employ a convex loss function $\ell$.
As for the fairness constraint we use the linear loss function, which implies the constraint to be convex.
Let $\boldsymbol{u}_g$ be the barycenter in the feature space of the positively labelled points in the group $g\in \{a,b\}$, that is
\begin{align}
\textstyle
\boldsymbol{u}_g= \frac{1}{n^{+,g}} \sum_{ i \in \mathcal{I}^{+,g}}
\boldsymbol{\phi}(\boldsymbol{x}_i),
\end{align}
where $\mathcal{I}^{+,g} = \{i: y_i {=} 1, x_{i,1} {=} g \}$.
Then using Eq.~\eqref{eq:222} the constraint in Problem~\eqref{eq:problemSoft} takes the form $\big|\langle \boldsymbol{w},\boldsymbol{u}_a-\boldsymbol{u}_b\rangle\big| \leq \epsilon$.

In practice, we solve the Tikhonov regularization problem
\begin{align}
\textstyle
\min\limits_{\boldsymbol{w} \in \mathbb{H}} \ 
\sum_{i =1}^n \ell(\langle \boldsymbol{w},\boldsymbol{\phi}(\boldsymbol{x}_i)\rangle ,y_i) + \lambda \|\boldsymbol{w}\|^2 \quad 
\text{s.t.}\ \big|\langle \boldsymbol{w},\boldsymbol{u}\rangle\big| \leq \epsilon 
\label{prob:ker} 
\end{align}
where $\boldsymbol{u} = \boldsymbol{u}_a - \boldsymbol{u}_b$ and $\lambda$ is a positive parameter which controls model complexity.
In particular, if $\epsilon = 0$ the constraint in Problem~\eqref{prob:ker} reduces to an orthogonality constraint that has a simple geometric interpretation.
Specifically, the vector $\boldsymbol{w}$ is required to be orthogonal to the vector formed by the difference between the barycenters of the positive labelled input samples in the two groups.

By the representer theorem~\cite{scholkopf2001generalized}, the solution to Problem~\eqref{prob:ker} is a linear combination of the feature vectors $\boldsymbol{\phi}(\boldsymbol{x}_1),\dots,\boldsymbol{\phi}(\boldsymbol{x}_n)$ and the vector $\boldsymbol{u}$.
However, in our case $\boldsymbol{ u}$ is itself a linear combination of the feature vectors (in fact only those corresponding to the subset of positive labeled points) hence $\boldsymbol{w}$ is a linear combination of the input points, that is $\boldsymbol{ w}=\sum_{i=1}^n \alpha_i\phi(\boldsymbol{x}_i)$.
The corresponding function used to make predictions is then given by $f(\boldsymbol{x}) = \sum_{i=1}^n \alpha_i \kappa(\boldsymbol{x}_i,\boldsymbol{x})$. 
Let $K$ be the Gram matrix.
The vector of coefficients $\boldsymbol{\alpha}$ can then be found by solving 
\begin{align}
\min_{\boldsymbol{\alpha} \in \mathbb{R}^n} \hspace{-.04truecm}\Bigg\{\hspace{-.04truecm} \sum_{i=1}^n \ell\bigg(\sum_{j=1}^n K_{ij}\alpha_j,y_i\bigg) {+} \lambda \! \! \sum_{i,j=1}^n\alpha_i \alpha_j K_{ij} \quad \hspace{-.04truecm}
\text{s.t.} \ & 
\bigg| 
\hspace{-.04truecm}\sum_{i=1}^n \alpha_i
\bigg[ 
\frac{1}{n^{+,a}} \!\!\! \hspace{-.04truecm}\sum_{j \in \mathcal{I}^{+,a}} \!\!\! K_{ij} 
{-} 
\frac{1}{n^{+,b}} \!\!\! \sum_{j \in \mathcal{I}^{+,b}} \!\!\! K_{ij} 
\bigg] 
\bigg| \leq \epsilon\Bigg\}.
\nonumber
\end{align}
In our experiments below we consider this particular case of Problem~\eqref{prob:ker} and furthermore choose the loss function $\ell_c$ to be the Hinge loss.
The resulting method is an extension of SVM.
The fairness constraint and, in particular, the orthogonality constraint when $\epsilon = 0$, can be easily added within standard SVM solvers\footnote{In supplementary material we derive the dual of Problem~\eqref{prob:ker} when $\ell_c$ is the Hinge loss.}

It is instructive to consider Problem~\eqref{prob:ker} when $\boldsymbol{\phi}$ is the identity mapping (i.e.~$\kappa$ is the linear kernel on $\mathbb{R}^d$) and $\epsilon=0$.
In this special case we can solve the orthogonality constraint $\langle \boldsymbol{w},\boldsymbol{u}\rangle = 0$ for $w_i$, where the 
index $i$ is such that $| u_i | = \|\boldsymbol{u}\|_\infty$, obtaining that $w_{i} = - \sum_{j=1, j \neq i}^d w_j \frac{u_j}{u_i}$.
Consequently the linear model rewrites as
$\sum_{j=1}^{d} w_j x_j = \sum_{j=1, j \neq i}^d w_j (x_j - x_i \frac{u_i}{u_j})$.
In this way, we then see the fairness constraint is implicitly enforced by making the change of representation $\boldsymbol{x} \mapsto \boldsymbol{\tilde{x}} \in \mathbb{R}^{d-1}$, with
\begin{equation}
\textstyle
\tilde{x}_j = x_j - x_i \frac{u_i}{u_j}, \quad j \in \{ 1, \dots, i-1, i+1, \dots, d \}.
\label{eq:gggg}
\end{equation}
In other words, we are able to obtain a fair linear model without any other constraint and by using a representation that has one feature fewer than the original one\footnote{
In supplementary material is reported the generalization of this argument to kernel for SVM.}
\section{Experiments}
\label{sec:exps}
In this section, we present numerical experiments with the proposed method on one synthetic and five real datasets.
The aim of the experiments is threefold.
First, we show that our approach is effective in selecting a fair model, incurring only a moderate loss in accuracy.
Second, we provide an empirical study of the properties of the method, which supports our theoretical observations in~ Section~\ref{sec:luca:th:Fairness}.
Third, we highlight the generality of our approach by showing that it can be used effectively within other linear models such as Lasso.
 
We use our approach with $\epsilon {=} 0$ in order to simplify the hyperparameter selection procedure.
For the sake of completeness, a set of results for different values of $\epsilon$ is presented in the supplementary material and briefly we comment on these below.
In all the experiments, we collect statistics concerning the classification accuracy and DEO of the selected model.
We recall that the DEO is defined in Eq.~\eqref{def:DEOmichele} and is the absolute difference of the true positive rate of the classifier applied to the two groups.
In all experiments, we performed a 10-fold cross validation (CV) to select the best hyperparameters\footnote{The regularization parameter $C$ (for both SVM and our method) with $30$ values, equally spaced in logarithmic scale 
between $10^{-4}$ and $10^{4}$; we used both the linear or RBF kernel (i.e.~for two examples $\boldsymbol{x}$ and $\boldsymbol{z}$, the RBF kernel is $e^{-\gamma ||\boldsymbol{x}-\boldsymbol{z}||^2}$) with $\gamma \in \{0.001, 0.01, 0.1, 1\}$.
In our case, $C=\frac{1}{2 \lambda}$ of Eq.~\eqref{prob:ker}.}.
For the Arrhythmia, COMPAS, German and Drug datasets, this procedure is repeated $10$ times, and we reported the average performance on the test set alongside its standard deviation.
For the Adult dataset, we used the provided split of train and test sets.
Unless otherwise stated, we employ two steps in the 10-fold CV procedure.
In the first step, the value of the hyperparameters with highest accuracy is identified.
In the second step, we shortlist all the hyperparameters with accuracy close to the best one (in our case, above $90 \%$ of the best accuracy).
Finally, from this list, we select the hyperparameters with the lowest DEO.
This novel validation procedure, that we wil call NVP, is a sanity-check to ensure that fairness cannot be achieved by a simple modification of hyperparameter selection procedure.
{\bf The code of our method is available at:} \url{https://github.com/jmikko/fair_ERM}.

\noindent \textbf{Synthetic Experiment.}
The aim of this experiment is to study the behavior of our method, in terms of both DEO and classification accuracy, in comparison to standard SVM (with our novel validation procedure).
To this end, we generated a synthetic binary classification dataset with two sensitive groups in the following manner.
For each group in the class $-1$ and for the group $a$ in the class $+1$, we generated $1000$ examples for training and the same amount for testing.
For the group $b$ in the class $+1$, we generated $200$ examples for training and the same number for testing.
Each set of examples is sampled from a $2$-dimensional isotropic Gaussian distribution with different mean $\mu$ and variance $\sigma^2$: (i) Group $a$, Label $+1$: $\mu=(-1, -1)$, $\sigma^2=0.8$; (ii) Group $a$, Label $-1$: $\mu=(1, 1)$, $\sigma^2=0.8$; (iii) Group $b$, Label $+1$: $\mu=(0.5, -0.5)$, $\sigma^2=0.5$; (iv) Group $b$, Label $-1$: $\mu=(0.5, 0.5)$, $\sigma^2=0.5$.
When a standard machine learning method is applied to this toy dataset, the generated model is unfair with respect to the group $b$, in that the classifier tends to negatively classify the examples in this group.

\begin{figure}
\centering
\includegraphics[trim={0.2cm 0.2cm 1.5cm 0.0cm},clip,width=.29\columnwidth]{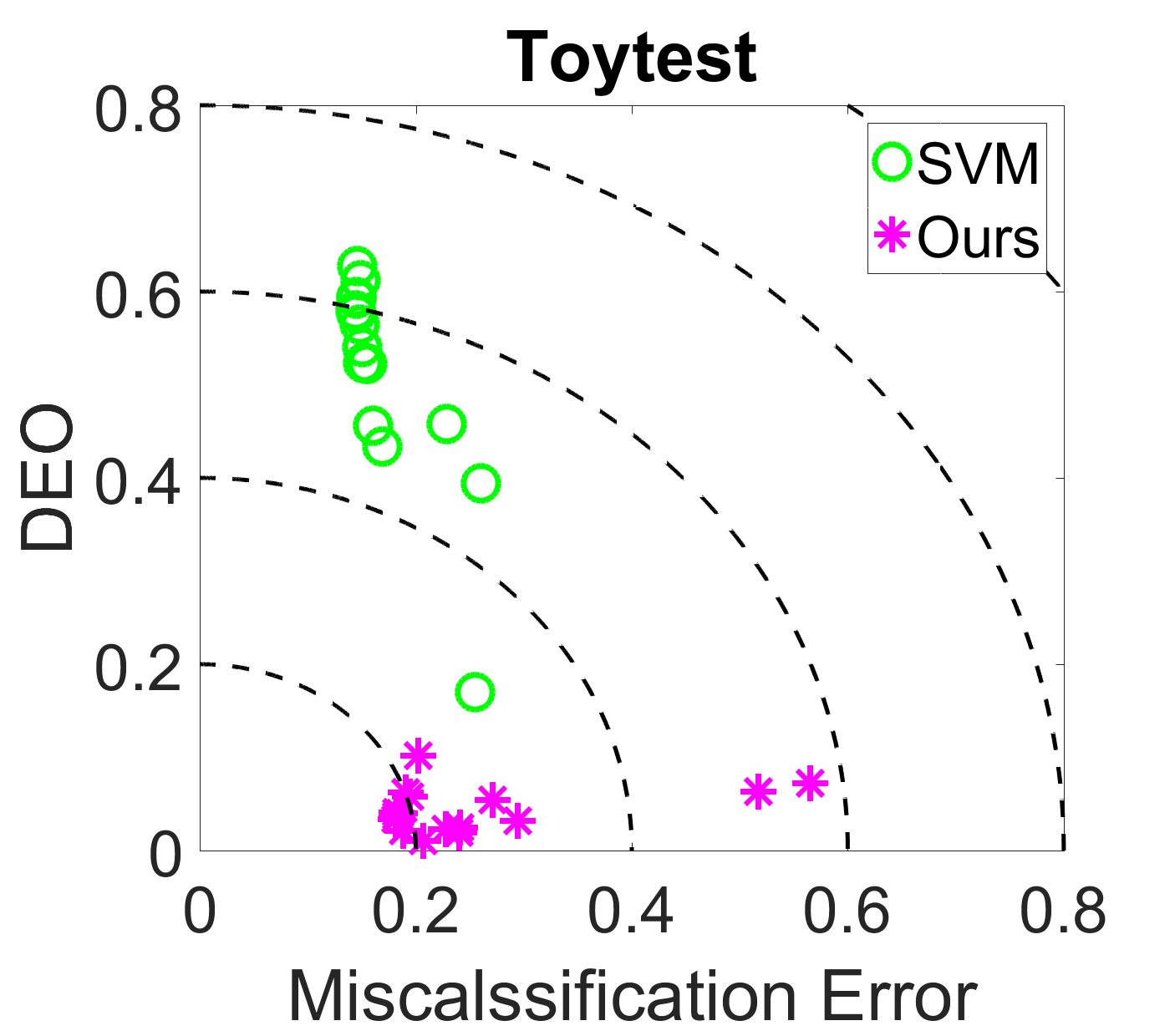}\hspace{.5truecm}
\includegraphics[trim={0.8cm 0.5cm 1.5cm 0.8cm},clip,width=.31\columnwidth]{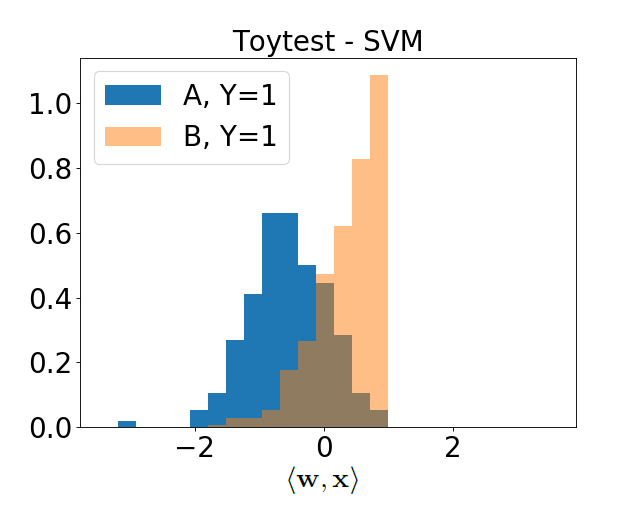} \hspace{.5truecm} 
\includegraphics[trim={2.1cm 0.5cm 1.5cm 0.8cm},clip,width=.285\columnwidth]{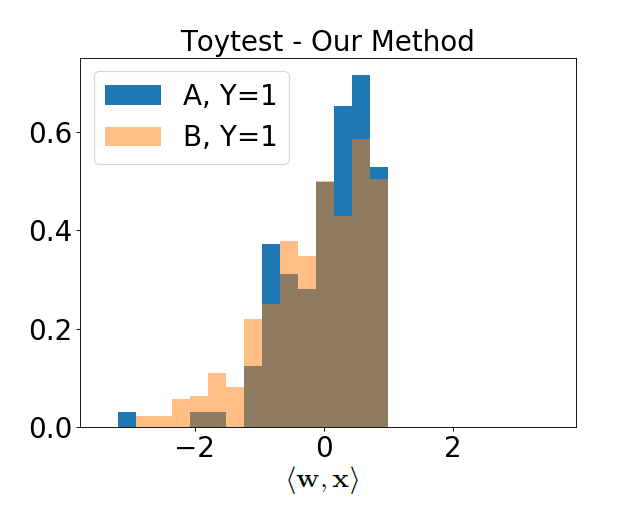}
\caption{Left: Test classification error and DEO for different values of hyperparameter $C$ for standard linear SVM (green circles) and our modified linear SVM (magenta stars).
Center and Rights: Histograms of the distribution of the values $\langle \boldsymbol{w}, \boldsymbol{x}\rangle$ for the two groups ($a$ in blue and $b$ in light orange) for test examples with label equals to $+1$.
The results are collected by using the optimal validated model for the classical linear SVM (Center) and for our linear method (Right).}
\label{fig:toydistrib}
\end{figure}

We trained different models, varying the value of the hyperparameter $C$, and using the standard linear SVM and our linear method.
Figure~\ref{fig:toydistrib} (Left) shows the performance of the various generated models with respect to the classification error and DEO on the test set.
Note that our method generated models that have an higher level of fairness, maintaining a good level of accuracy.
The grid in the plots emphasizes the fact that both the error and DEO have to be simultaneously considered in the evaluation of a method.
%
Figure~\ref{fig:toydistrib} (Center and Left) depicts the histogram of the values of $\langle \boldsymbol{w}, \boldsymbol{x}\rangle $ (where $\boldsymbol{w}$ is the generated model) for test examples with true label equal to $+1$ for each of the two groups.
The results are reported both for our method (Right) and standard SVM (Center).
Note that our method generates a model with a similar true positive rate among the two groups (i.e.~the areas of the value when the horizontal axis is greater than zero are similar for groups $a$ and $b$).
Moreover, due to the simplicity of the toy test, the distribution with respect to the two different groups is also very similar when our model is used.
%
%

\noindent \textbf{Real Data Experiments.}
We next compare the performance of our model to set of different methods on $5$ publicly available datasets:
Arrhythmia, COMPAS, Adult,
German, and Drug.
A description of the datasets is provided in the supplementary material.
These datasets have been selected from the standard databases of datasets (UCI, mldata and Fairness-Measures\footnote{Fairness-Measures website: \url{fairness-measures.org}}).
We considered only datasets with a DEO higher than $0.1$, when the model is generated by an SVM validated with the NVP.
For this reason, some of the commonly used datasets have been discarded (e.g.~Diabetes, Heart, SAT, PSU-Chile, and SOEP).
We compared our method both in the linear and not linear case against: (i) Na\"{i}ve SVM, validated with a standard nested 10-fold CV procedure.
This method ignores fairness in the validation procedure, simply trying to optimize accuracy; (ii) SVM with the NVP.
As noted above, this baseline is the simplest way to inject the fairness into the model; (iii) Hardt method~\cite{hardt2016equality} applied to the best SVM; (iv) Zafar method~\cite{zafar2017fairness}, implemented with the code provided by the authors for the linear case\footnote{Python code for~\cite{zafar2017fairness}: \url{https://github.com/mbilalzafar/fair-classification}}.
Concerning our method, in the linear case, it exploits the preprocessing presented in Section~\ref{sec:luca:th:FK}.
\begin{table*}
\tiny
\centering
\setlength{\tabcolsep}{0.05cm}
\begin{tabular}{|l|c|c|c|c|c|c|c|c|c|c|c|c|c|c|c|c|c|c|c|c|}
\hline
\hline
& \multicolumn{2}{c|}{Arrhythmia} 
& \multicolumn{2}{c|}{COMPAS} 	
& \multicolumn{2}{c|}{Adult} 
& \multicolumn{2}{c|}{German} 
& \multicolumn{2}{c|}{Drug} \\
Method & ACC & DEO & ACC & DEO & ACC & DEO & ACC & DEO & ACC & DEO \\
\hline
\hline 
\multicolumn{11}{c}{$s$ not inside $\boldsymbol{x}$}\\
\hline
\hline
Na\"{i}ve Lin. SVM & $0.75 {\pm} 0.04$ & $0.11 {\pm} 0.03$ & $0.73 {\pm} 0.01$	& $0.13 {\pm} 0.02$ & $0.78$ & $0.10$ & $0.71 {\pm} 0.06$ & $0.16 {\pm} 0.04$ & $0.79 {\pm} 0.02$ & $0.25 {\pm} 0.03$ \\
Lin. SVM & $0.71 {\pm} 0.05$	& $0.10 {\pm} 0.03$ & $0.72 {\pm} 0.01$	& $0.12 {\pm} 0.02$ & $0.78$ & $0.09$ & $0.69 {\pm} 0.04$ & $0.11 {\pm} 0.10$ & $0.79 {\pm} 0.02$ & $0.25 {\pm} 0.04$	\\
Hardt & - & - & - & - & - & - & - & - & -	& -			\\
Zafar & $0.67 {\pm} 0.03$ & $0.05 {\pm} 0.02$	& $0.69 {\pm} 0.01$ & $0.10 {\pm} 0.08$ & $0.76$ & $0.05$ & $0.62 {\pm} 0.09$ & $0.13 {\pm} 0.10$ & $0.66 {\pm} 0.03$ & $0.06 {\pm} 0.06$		\\
Lin. Ours & $0.75 {\pm} 0.05$ & $0.05 {\pm} 0.02$	& $0.73 {\pm} 0.01$	 & $0.07 {\pm} 0.02$ & $0.75$ & $0.01$ & $0.69 {\pm} 0.04$ & $0.06 {\pm} 0.03$ & $0.79 {\pm} 0.02$ & $0.10 {\pm} 0.06$ \\
\hline
Na\"{i}ve SVM & $0.75 {\pm} 0.04$ & $0.11 {\pm} 0.03$ & $0.72 {\pm} 0.01$	& $0.14 {\pm} 0.02$ & $0.80$ & $0.09$ & $0.74 {\pm} 0.05$ & $0.12 {\pm} 0.05$ & $0.81 {\pm} 0.02$ & $0.22 {\pm} 0.04$	\\
SVM & $0.71 {\pm} 0.05$	& $0.10 {\pm} 0.03$ & $0.73 {\pm} 0.01$	& $0.11 {\pm} 0.02$ & $0.79$ & $0.08$ & $0.74 {\pm} 0.03$ & $0.10 {\pm} 0.06$ & $0.81 {\pm} 0.02$ & $0.22 {\pm} 0.03$ 			\\
Hardt & - & - & - & - & - & - & - & - & -	& -	 				\\
Ours & $0.75 {\pm} 0.05$ & $0.05 {\pm} 0.02$ & $0.72 {\pm} 0.01$	& $0.08 {\pm} 0.02$ & $0.77$ & $0.01$ & $0.73 {\pm} 0.04$ & $0.05 {\pm} 0.03$ & $0.79 {\pm} 0.03$ & $0.10 {\pm} 0.05$			\\
\hline
\hline 
\multicolumn{11}{c}{$s$ inside $\boldsymbol{x}$}\\
\hline
\hline
Na\"{i}ve Lin. SVM 	& $0.79 {\pm} 0.06$ & $0.14 {\pm} 0.03$	& $0.76 {\pm} 0.01$ & $0.17 {\pm} 0.02$ & $0.81$ & $0.14$ & $0.71 {\pm} 0.06$ & $0.17 {\pm} 0.05$ & $0.81 {\pm} 0.02$ & $0.44 {\pm} 0.03$ \\
Lin. SVM 				& $0.78 {\pm} 0.07$ & $0.13 {\pm} 0.04$& $0.75 {\pm} 0.01$ & $0.15 {\pm} 0.02$ & $0.80$ & $0.13$ & $0.69 {\pm} 0.04$ & $0.11 {\pm} 0.10$ & $0.81 {\pm} 0.02$ & $0.41 {\pm} 0.06$ \\
Hardt 					& $0.74 {\pm} 0.06$ & $0.07 {\pm} 0.04$& $0.67 {\pm} 0.03$ & $0.21 {\pm} 0.09$ & $0.80$ & $0.10$ & $0.61 {\pm} 0.15$ & $0.15 {\pm} 0.13$ & $0.77 {\pm} 0.02$ & $0.22 {\pm} 0.09$ \\
Zafar 					& $0.71 {\pm} 0.03$ & $0.03 {\pm} 0.02$	& $0.69 {\pm} 0.02$ & $0.10 {\pm} 0.06$ & $0.78$ & $0.05$ & $0.62 {\pm} 0.09$ & $0.13 {\pm} 0.11$ & $0.69 {\pm} 0.03$ & $0.02 {\pm} 0.07$ \\
Lin. Ours 			& $0.79 {\pm} 0.07$ & $0.04 {\pm} 0.03$	& $0.76 {\pm} 0.01$ & $0.04 {\pm} 0.03$ & $0.77$ & $0.01$ & $0.69 {\pm} 0.04$ & $0.05 {\pm} 0.03$ & $0.79 {\pm} 0.02$ & $0.05 {\pm} 0.03$ \\
\hline
Na\"{i}ve SVM 			& $0.79 {\pm} 0.06$ & $0.14 {\pm} 0.04$& $0.76 {\pm} 0.01$ & $0.18 {\pm} 0.02$ & $0.84$ & $0.18$ & $0.74 {\pm} 0.05$ & $0.12 {\pm} 0.05$ & $0.82 {\pm} 0.02$ & $0.45 {\pm} 0.04$ \\
SVM 					& $0.78 {\pm} 0.06$ & $0.13 {\pm} 0.04$& $0.73 {\pm} 0.01$ & $0.14 {\pm} 0.02$ & $0.82$ & $0.14$ & $0.74 {\pm} 0.03$ & $0.10 {\pm} 0.06$ & $0.81 {\pm} 0.02$ & $0.38 {\pm} 0.03$ \\
Hardt 					& $0.74 {\pm} 0.06$ & $0.07 {\pm} 0.04$	& $0.71 {\pm} 0.01$ & $0.08 {\pm} 0.01$ & $0.82$ & $0.11$ & $0.71 {\pm} 0.03$ & $0.11 {\pm} 0.18$ & $0.75 {\pm} 0.11$ & $0.14 {\pm} 0.08$ \\
Ours 					& $0.79 {\pm} 0.09$ & $0.03 {\pm} 0.02$& $0.73 {\pm} 0.01$ & $0.05 {\pm} 0.03$ & $0.81$ & $0.01$ & $0.73 {\pm} 0.04$ & $0.05 {\pm} 0.03$ & $0.80 {\pm} 0.03$ & $0.07 {\pm} 0.05$ \\ 
\hline
\hline
\end{tabular}
\caption{Results (average $\pm$ standard deviation, when a fixed test set is not provided) for all the datasets, concerning accuracy (ACC) and DEO \text.}
\vspace{-.5cm}
\label{tab:results}
\end{table*}

\begin{figure}
\centering
\includegraphics[width=0.45\columnwidth]{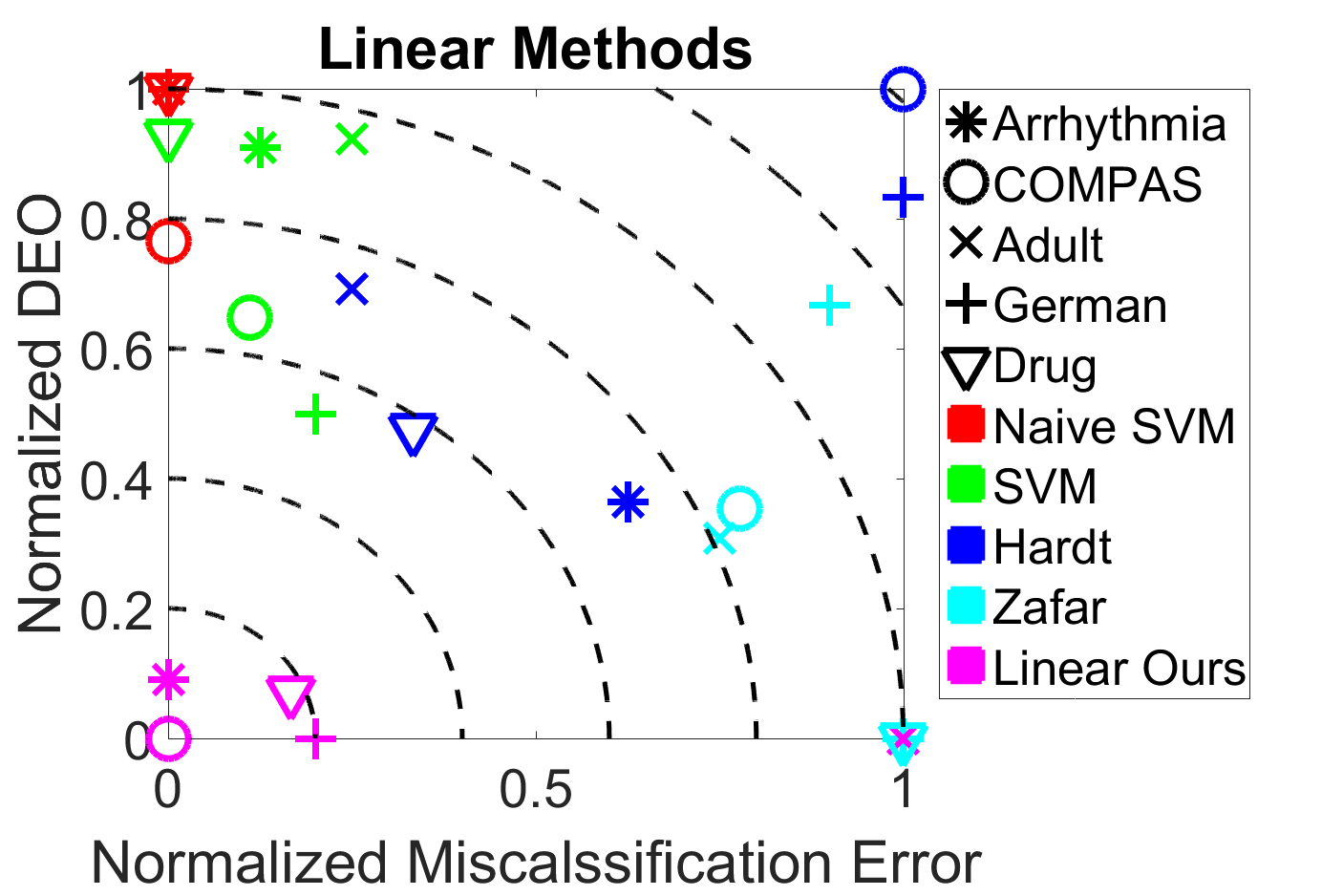} \quad
\includegraphics[width=0.45\columnwidth]{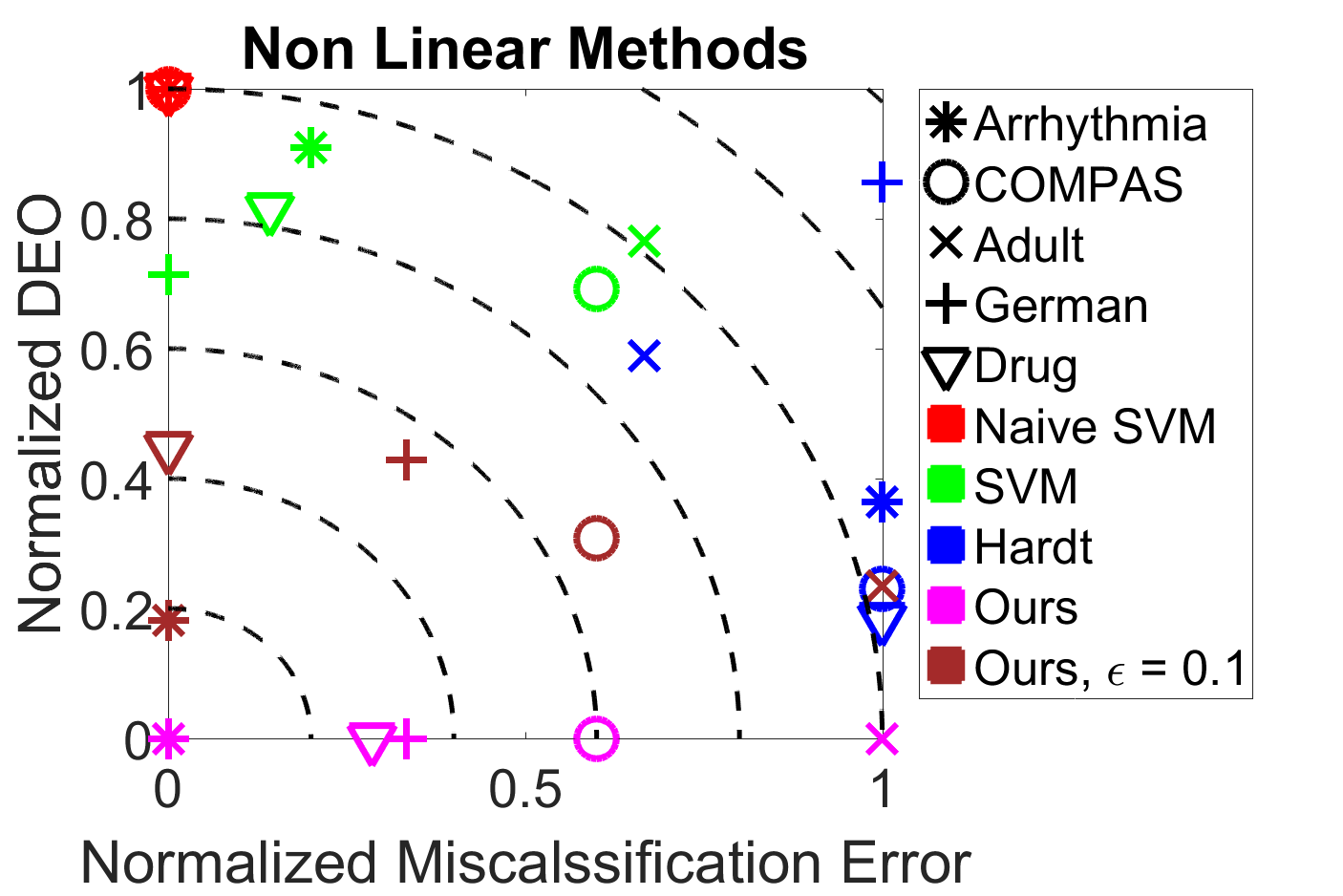}
\caption{{Results of Table~\ref{tab:results} of linear (left) and nonlinear (right) methods, when the error and the DEO are normalized in $[0,1]$ column-wise and when $s$ is inside $\boldsymbol{x}$. Different symbols and colors refer to different datasets and method respectively. The closer a point is to the origin, the better the result is.}}
\vspace{-.4cm}
\label{fig:tableexplanation}
\end{figure}

Table~\ref{tab:results} shows our experimental results for all the datasets and methods both when $s$ is inside $\boldsymbol{x}$ or not.
This result suggests that our method performs favorably over the competitors in that it decreases DEO substantially with only a moderate loss in accuracy.
Moreover having $s$ inside $\boldsymbol{x}$ increases the accuracy but - for the methods without the specific purpose of producing fairness models - decreases the fairness. On the other hand, having $s$ inside $\boldsymbol{x}$ ensures to our method the ability of improve the fairness by exploiting the value of $s$ also in the prediction phase. This is to be expected, since knowing the group membership increases our information but also leads to behaviours able to influence the fairness of the predictive model.
In order to quantify this effect, we present in Figure~\ref{fig:tableexplanation} the results of Table~\ref{tab:results} of linear (left) and nonlinear (right) methods, when the error (one minus accuracy) and the DEO are normalized in $[0,1]$ column-wise and when the $s$ is inside $\boldsymbol{x}$\footnote{The case when $s$ is not inside $\boldsymbol{x}$ is reported in the supplementary materials).}.
In the figure, different symbols and colors refer to different datasets and methods, respectively.
The closer a point is to the origin, the better the result is.
The best accuracy is, in general, reached by using the Na\"{i}ve SVM (in red) both for the linear and nonlinear case.
This behavior is expected due to the absence of any fairness constraint.
On the other hand, Na\"{i}ve SVM has unsatisfactory levels of fairness.
Hardt~\cite{hardt2016equality} (in blue) and Zafar~\cite{zafar2017fairness} (in cyan, for the linear case) methods are able to obtain a good level of fairness but the price of this fair model is a strong decrease in accuracy.
Our method (in magenta) obtains similar or better results concerning the DEO preserving the performance in accuracy.
In particular in the nonlinear case, our method reaches the lowest levels of DEO with respect to all the methods.
For the sake of completeness, in the nonlinear (bottom) part of Figure~\ref{fig:tableexplanation}, we show our method when the parameter $\epsilon$ is set to $0.1$ (in brown) instead of $0$ (in magenta).
As expected, the generated models are less fair with a (small) improvement in the accuracy.
An in depth analysis of the role of $\epsilon$ is presented in supplementary materials.

\noindent \textbf{Application to Lasso.} 
%
Due to the particular proposed methodology, we are able in principle to apply our method to any learning algorithm. In particular, when the algorithm generates a linear model we can exploit the data preprocessing in 
Eq.~\eqref{eq:gggg}, to directly impose fairness in the model. Here, we show how it is possible to obtain a sparse and fair model by exploiting the standard Lasso algorithm in synergy with this preprocessing step. For this purpose, we selected the Arrhythmia dataset as the Lasso works well in a high dimensional / small sample setting.
We performed the same experiment described above,
where we used the Lasso algorithm in place of the SVM. 
In this case, by Na\"{i}ve Lasso, we refer to the Lasso when it is validated with a standard nested 10-fold CV procedure, whereas by Lasso we refer to the standard Lasso with the NVP outlined above. 
The method of~\cite{hardt2016equality} has been applied to the best Lasso model. 
Moreover, we reported the results obtained using Na\"{i}ve Linear SVM and Linear SVM. We also repeated the experiment by using a reduced training set in order to highlight the effect of the sparsity. 
Table~\ref{tab:results_lasso} reported in the supplementary material shows the results. It is possible to note that, reducing the training sets, the generated models become less fair (i.e. the DEO increases). Using our method, we are able to maintain a fair model reaching satisfactory accuracy results. 

\noindent \textbf{The Value of $\hat{\Delta}$.} 
Finally, we show experimental results to highlight how the hypothesis of Proposition~\ref{thm:mainresult2} (Section~\ref{sec:luca:th:FERM}) are reasonable in the real cases. We know that, if the hypothesis of inequality (\ref{eq:hp1}) are satisfied, the linear loss based fairness is close to the EO. Specifically, these two quantities are closer when $\hat{\Delta}$ is small. We evaluated $\hat{\Delta}$ for benchmark and toy datasets. The obtained results are in Table~\ref{tab:delta} of supplementary material, where $\hat{\Delta}$ has the order of magnitude of $10^{-2}$ in all the datasets. Consequently, our method is able to obtain a good approximation of the DEO.
\vspace{-.3cm}
\section{Conclusion and Future Work}
\label{sec:conc}
\vspace{-.1cm}
We have presented a generalized notion of fairness, which encompasses previously introduced notion and can be used to constrain ERM, in order to learn fair classifiers. The framework is appealing both theoretically and practically. Our theoretical observations provide a statistical justification for this approach and our algorithmic observations suggest a way to implement it efficiently in the setting of kernel methods. Experimental results suggest that our approach is promising for applications, generating models with improved fairness properties while maintaining classification accuracy. We close by mentioning directions of future research. On the algorithmic side, it would be interesting to study whether our method can be improved by other relaxations of the fairness constraint beyond the linear loss used here. Applications of the fairness constraint to multi-class classification or to regression tasks would also be valuable. On the theory side, it would be interesting to study how the choice of the parameter $\epsilon$ affects the statistical performance of our method and derive optimal accuracy-fairness trade-off as a function of this parameter. 

\section*{Acknowledgments}
This work was supported by the Amazon AWS Machine Learning Research Award.

{
\small
\bibliographystyle{unsrt}
\bibliography{biblio}

\begin{thebibliography}{10}

\bibitem{dwork2018decoupled}
C.~Dwork, N.~Immorlica, A.~T. Kalai, and M.~D.~M. Leiserson.
\newblock Decoupled classifiers for group-fair and efficient machine learning.
\newblock In {\em Conference on Fairness, Accountability and Transparency},
  2018.

\bibitem{hardt2016equality}
M.~Hardt, E.~Price, and N.~Srebro.
\newblock Equality of opportunity in supervised learning.
\newblock In {\em Advances in neural information processing systems}, 2016.

\bibitem{zafar2017fairness}
M.~B. Zafar, I.~Valera, M.~Gomez~Rodriguez, and K.~P. Gummadi.
\newblock Fairness beyond disparate treatment \& disparate impact: Learning
  classification without disparate mistreatment.
\newblock In {\em International Conference on World Wide Web}, 2017.

\bibitem{zemel2013learning}
R.~Zemel, Y.~Wu, K.~Swersky, T.~Pitassi, and C.~Dwork.
\newblock Learning fair representations.
\newblock In {\em International Conference on Machine Learning}, 2013.

\bibitem{kilbertus2017avoiding}
N.~Kilbertus, M.~Rojas-Carulla, G.~Parascandolo, M.~Hardt, D.~Janzing, and
  B.~Sch{\"o}lkopf.
\newblock Avoiding discrimination through causal reasoning.
\newblock In {\em Advances in Neural Information Processing Systems}, 2017.

\bibitem{kusner2017counterfactual}
M.~J. Kusner, J.~Loftus, C.~Russell, and R.~Silva.
\newblock Counterfactual fairness.
\newblock In {\em Advances in Neural Information Processing Systems}, 2017.

\bibitem{calmon2017optimized}
F.~Calmon, D.~Wei, B.~Vinzamuri, K.~Natesan Ramamurthy, and K.~R. Varshney.
\newblock Optimized pre-processing for discrimination prevention.
\newblock In {\em Advances in Neural Information Processing Systems}, 2017.

\bibitem{joseph2016fairness}
M.~Joseph, M.~Kearns, J.~H. Morgenstern, and A.~Roth.
\newblock Fairness in learning: Classic and contextual bandits.
\newblock In {\em Advances in Neural Information Processing Systems}, 2016.

\bibitem{chierichetti2017fair}
F.~Chierichetti, R.~Kumar, S.~Lattanzi, and S.~Vassilvitskii.
\newblock Fair clustering through fairlets.
\newblock In {\em Advances in Neural Information Processing Systems}, 2017.

\bibitem{jabbari2016fair}
S.~Jabbari, M.~Joseph, M.~Kearns, J.~Morgenstern, and A.~Roth.
\newblock Fair learning in markovian environments.
\newblock In {\em Conference on Fairness, Accountability, and Transparency in
  Machine Learning}, 2016.

\bibitem{yao2017beyond}
S.~Yao and B.~Huang.
\newblock Beyond parity: Fairness objectives for collaborative filtering.
\newblock In {\em Advances in Neural Information Processing Systems}, 2017.

\bibitem{lum2016statistical}
K.~Lum and J.~Johndrow.
\newblock A statistical framework for fair predictive algorithms.
\newblock {\em arXiv preprint arXiv:1610.08077}, 2016.

\bibitem{zliobaite2015relation}
I.~Zliobaite.
\newblock On the relation between accuracy and fairness in binary
  classification.
\newblock {\em arXiv preprint arXiv:1505.05723}, 2015.

\bibitem{calders2009building}
T.~Calders, F.~Kamiran, and M.~Pechenizkiy.
\newblock Building classifiers with independency constraints.
\newblock In {\em IEEE international conference on Data mining}, 2009.

\bibitem{pleiss2017fairness}
G.~Pleiss, M.~Raghavan, F.~Wu, J.~Kleinberg, and K.~Q. Weinberger.
\newblock On fairness and calibration.
\newblock In {\em Advances in Neural Information Processing Systems}, 2017.

\bibitem{beutel2017data}
A.~Beutel, J.~Chen, Z.~Zhao, and E.~H. Chi.
\newblock Data decisions and theoretical implications when adversarially
  learning fair representations.
\newblock In {\em Conference on Fairness, Accountability, and Transparency in
  Machine Learning}, 2017.

\bibitem{feldman2015certifying}
M.~Feldman, S.~A. Friedler, J.~Moeller, C.~Scheidegger, and
  S.~Venkatasubramanian.
\newblock Certifying and removing disparate impact.
\newblock In {\em International Conference on Knowledge Discovery and Data
  Mining}, 2015.

\bibitem{agarwal2017reductions}
A.~Agarwal, A.~Beygelzimer, M.~Dud{\'\i}k, and J.~Langford.
\newblock A reductions approach to fair classification.
\newblock In {\em Conference on Fairness, Accountability, and Transparency in
  Machine Learning}, 2017.

\bibitem{agarwal2018reductions}
A.~Agarwal, A.~Beygelzimer, M.~Dud{\'\i}k, J.~Langford, and H.~Wallach.
\newblock A reductions approach to fair classification.
\newblock {\em arXiv preprint arXiv:1803.02453}, 2018.

\bibitem{woodworth2017learning}
B.~Woodworth, S.~Gunasekar, M.~I. Ohannessian, and N.~Srebro.
\newblock Learning non-discriminatory predictors.
\newblock In {\em Computational Learning Theory}, 2017.

\bibitem{menon2018cost}
A.~K. Menon and R.~C. Williamson.
\newblock The cost of fairness in binary classification.
\newblock In {\em Conference on Fairness, Accountability and Transparency},
  2018.

\bibitem{zafar2017parity}
M.~B. Zafar, I.~Valera, M.~Rodriguez, K.~Gummadi, and A.~Weller.
\newblock From parity to preference-based notions of fairness in
  classification.
\newblock In {\em Advances in Neural Information Processing Systems}, 2017.

\bibitem{bechavod2018Penalizing}
Y.~Bechavod and K.~Ligett.
\newblock Penalizing unfairness in binary classification.
\newblock {\em arXiv preprint arXiv:1707.00044v3}, 2018.

\bibitem{zafar2017fairnessARXIV}
M.~B. Zafar, I.~Valera, M.~Gomez~Rodriguez, and K.~P. Gummadi.
\newblock Fairness constraints: Mechanisms for fair classification.
\newblock In {\em International Conference on Artificial Intelligence and
  Statistics}, 2017.

\bibitem{kamishima2011fairness}
T.~Kamishima, S.~Akaho, and J.~Sakuma.
\newblock Fairness-aware learning through regularization approach.
\newblock In {\em International Conference on Data Mining Workshops}, 2011.

\bibitem{kearns2017preventing}
M.~Kearns, S.~Neel, A.~Roth, and Z.~S. Wu.
\newblock Preventing fairness gerrymandering: Auditing and learning for
  subgroup fairness.
\newblock {\em arXiv preprint arXiv:1711.05144}, 2017.

\bibitem{Prez-Suay2017Fair}
A.~P{\'e}rez-Suay, V.~Laparra, G.~Mateo-Garc{\'\i}a, J.~Mu{\~{n}}oz-Mar{\'\i},
  L.~G{\'o}mez-Chova, and G.~Camps-Valls.
\newblock Fair kernel learning.
\newblock In {\em Machine Learning and Knowledge Discovery in Databases}, 2017.

\bibitem{berk2017convex}
R.~Berk, H.~Heidari, S.~Jabbari, M.~Joseph, M.~Kearns, J.~Morgenstern, S.~Neel,
  and A.~Roth.
\newblock A convex framework for fair regression.
\newblock {\em arXiv preprint arXiv:1706.02409}, 2017.

\bibitem{alabi2018optimizing}
D.~Alabi, N.~Immorlica, and A.~T. Kalai.
\newblock When optimizing nonlinear objectives is no harder than linear
  objectives.
\newblock {\em arXiv preprint arXiv:1804.04503}, 2018.

\bibitem{olfat2018spectral}
M.~Olfat and A.~Aswani.
\newblock Spectral algorithms for computing fair support vector machines.
\newblock In {\em International Conference on Artificial Intelligence and
  Statistics}, 2018.

\bibitem{adebayo2016iterative}
J.~Adebayo and L.~Kagal.
\newblock Iterative orthogonal feature projection for diagnosing bias in
  black-box models.
\newblock In {\em Conference on Fairness, Accountability, and Transparency in
  Machine Learning}, 2016.

\bibitem{kamiran2009classifying}
F.~Kamiran and T.~Calders.
\newblock Classifying without discriminating.
\newblock In {\em International Conference on Computer, Control and
  Communication}, 2009.

\bibitem{kamiran2012data}
F.~Kamiran and T.~Calders.
\newblock Data preprocessing techniques for classification without
  discrimination.
\newblock {\em Knowledge and Information Systems}, 33(1):1--33, 2012.

\bibitem{kamiran2010classification}
F.~Kamiran and T.~Calders.
\newblock Classification with no discrimination by preferential sampling.
\newblock In {\em Machine Learning Conference}, 2010.

\bibitem{shalev2014understanding}
S.~Shalev-Shwartz and S.~Ben-David.
\newblock {\em Understanding machine learning: From theory to algorithms}.
\newblock Cambridge University Press, 2014.

\bibitem{bartlett2002rademacher}
P.~L. Bartlett and S.~Mendelson.
\newblock Rademacher and gaussian complexities: Risk bounds and structural
  results.
\newblock {\em Journal of Machine Learning Research}, 3(Nov):463--482, 2002.

\bibitem{maurer2004note}
A.~Maurer.
\newblock A note on the pac bayesian theorem.
\newblock {\em arXiv preprint cs/0411099}, 2004.

\bibitem{shawe2004kernel}
J.~Shawe-Taylor and N.~Cristianini.
\newblock {\em Kernel methods for pattern analysis}.
\newblock Cambridge University Press, 2004.

\bibitem{smola2001}
A.~J. Smola and B.~Sch{\"o}lkopf.
\newblock {\em Learning with Kernels}.
\newblock MIT Press, 2001.

\bibitem{scholkopf2001generalized}
B.~Sch{\"o}lkopf, R.~Herbrich, and A.~Smola.
\newblock A generalized representer theorem.
\newblock In {\em Computational Learning Theory}, 2001.

\bibitem{vapnik1998statistical}
V.~N. Vapnik.
\newblock {\em Statistical learning theory}.
\newblock Wiley New York, 1998.

\bibitem{Rockafellar1970}
R.~T. Rockafellar.
\newblock {\em Convex Analysis}.
\newblock Princeton University Press, 1970.

\end{thebibliography}
}

\newpage

\section*{Supplementary Material}
\appendix
\section{Proofs}
\label{sec:SMproofs}
\begin{proof} [Proof of Theorem~\ref{thm:mainresult1}]
We first use Eq.~\eqref{eq:bartlett} to conclude that, with probability at least $1- 2 \delta$,
\begin{align}
\textstyle
\sup_{f \in \mathcal{F}} \Big| \big| L^{+,a}(f) - L^{+,b}(f) \big| - \big| \hat{L}^{+,a}(f) - \hat{L}^{+,b}(f) \big| \Big| \leq \sum\limits_{g \in \{a,b\}} B(\delta,n^{+,g},\mathcal{F}). 
\label{eq:proof1_eq2}
\end{align}
This inequality in turn implies that, with probability at least $1-2\delta$, it holds that
\begin{align}
\left\{ f: f \in \mathcal{F}, \big| L^{+,a}(f) - L^{+,b}(f) \big| \leq \epsilon \right\} 
\subseteq \left\{ f: f \in \mathcal{F}, \big| \hat{L}^{+,a}(f) - \hat{L}^{+,b}(f) \big| \leq \hat{\epsilon} \right\}.
\label{eq:proof1_eq3}
\end{align}
Now, in order to prove the first statement of the theorem, let us decompose the excess risk as
\begin{align}
L(\hat{f}) {-} L(f^*\!) = 
L(\hat{f}) 
- \hat{L}(\hat{f})
+ \hat{L}(\hat{f})
- \hat{L}(f^*\!) 
+ \hat{L}(f^*\!)
- L(f^*\!). \nonumber 
\end{align}
Inequality (\ref{eq:proof1_eq3}) implies that $\hat{L}(\hat{f}) - \hat{L}(f^*) \leq 0$ with probability at least $1 -2\delta$ and consequently with probability at least $1 - 2\delta$ it holds that \begin{align}
L(\hat{f}) - L(f^*)
\leq L(\hat{f}) - \hat{L}(\hat{f}) + 
\hat{L}(f^*) - L(f^*).
\nonumber 
\end{align}
The first statement now follows by Eq.~\eqref{eq:bartlett}. As for the second statement, its proof consists in exploiting the results of Eqns.~\eqref{eq:proof1_eq2} and~\eqref{eq:proof1_eq3} together with a union bound.
\end{proof}

\begin{proof}[Proof of Proposition~\ref{thm:mainresult2}]
The proof of the first statement follows directly by the inequality $\ell_h(f(\boldsymbol{x}),y) \leq \ell_c(f(\boldsymbol{x}),y)$. In order to prove the second statement, we first note that
\begin{align}
\textstyle
\big| \hat{L}^{+,a}_l(f) - \hat{L}^{+,b}_l(f)\big| = 
\frac{1}{2} \left| \hat{\mathbb{E}} \left[ f(\boldsymbol{x}) | y=1,s=a \right] - \hat{\mathbb{E}} \left[ f(\boldsymbol{x}) | y=1,s=b \right] \right|. \nonumber
\end{align}
By applying the same reasoning to $\big| \hat{L}^{+,a}_h(f) {-} \hat{L}^{+,b}_h(f)\big| $ and by exploiting inequality (\ref{eq:hp1}) the result follows.
\end{proof}

\section{Literature Review of Fairness Methods}
\label{sec:appreview}
In this section, we provide a brief analysis of the different existing methods concerning fairness. We show our findings in Table~\ref{tab:review}, where the rows represent properties, characteristics and experimental results of different fairness methods. The columns represent the different algorithms and, specifically, the first column is our approach. We think that, at this stage of development of fairness in machine learning, a clear understanding of the differences and similarities among the current available algorithms is a fundamental step. Table~\ref{tab:review} describes, in the first row, the family of the different methods, following the taxonomy defined in this paper (see Section~\ref{sec:intro}). The following $8$ rows describe general properties of the methods, as for example the convexity of the approach, the convergence of the learning phase or the consistency with respect to the risk and the fairness notion. The next $9$ rows describes the presence of a specific comparison between methods and, finally, in the last row the availability of the code online is analyzed.
\begin{table}[!ht]
\tiny
\centering
\setlength{\tabcolsep}{0.05cm}
\begin{tabular}{||l||c||c|c|c|c|c|c|c|c|c|c|c|c|c|c|c|c|c|c|c|c|c|c|c||}
\hline
\hline
\textbf{Ref.} & \textbf{Ours} & \cite{adebayo2016iterative} & \cite{calmon2017optimized} & \cite{agarwal2017reductions,agarwal2018reductions} & \cite{woodworth2017learning} & \cite{zafar2017fairness} & \cite{kamiran2009classifying} & \cite{Prez-Suay2017Fair} & \cite{zemel2013learning} & \cite{menon2018cost} & \cite{dwork2018decoupled} & \cite{zafar2017parity} & \cite{pleiss2017fairness} & \cite{beutel2017data} & \cite{bechavod2018Penalizing} & \cite{hardt2016equality} & \cite{zafar2017fairnessARXIV} & \cite{berk2017convex} & \cite{kamishima2011fairness} & \cite{feldman2015certifying} & 
\cite{kamiran2012data} & \cite{kamiran2010classification} &
\cite{alabi2018optimizing} & \cite{olfat2018spectral}\\
\hline
\hline
Method Family 
& 2\&3 & 3 & 3 & 2 & 2 & 2 & 3 & 2 & 3 & 2 & 2 & 2 & 1 & - & 2 & 1 & 2 & 2 & 2 & 1 & 3 & 3 & 2 & 2\\ 
\hline
\hline
Classification
& x & x & x & x & x & x & x & & x & x & x & x & x & x & x & x & x & & x & x & x & x & x & x \\
\hline
New Fairness Notions 
& x & x & x & x & x & x & x & x & x & x & x & x & x & & x & x & x & x & x & x & x & x & x & \\
\hline
Use of EO 
& x & & & x & & & & & & x & x & x & & & & x & & & & & & & & x\\
\hline
Convex Approach 
& x & x & & & x$^{*}$ & & & x & & & x & & & & x & & x & x & & & x & & x & x\\ 
\hline
Convergence Learning 
& x & x & x & x & & & & x & & & x & & & & & & & x & & & & & x &\\
\hline
Consistency Risk-Fairness 
& x & & & x & x & & & & & x & x & & & & & & & & & & & & &\\
\hline
Experimental Results 
& x & x & x & x & & x & x & x & x & x & x & x & x & x & x & x & x & x & x & x & x & x & & x\\
\hline
Epsilon validate 
& x & & & & & & & & & & & & & & & & & & & & & & &\\
\hline
Exp. w.r.t. \cite{hardt2016equality}
& x & & & x & & x & & & & & & & x & & x & & & & & & & & &\\
\hline
Exp. w.r.t. \cite{zafar2017fairness} 
& x & & & & & & & & & & & & x & & x & & & & & & & & & x\\
\hline
Exp. w.r.t. \cite{kamiran2012data}
& & & & x & & & & & & & & & & & & & & & & & & & &\\
\hline
Exp. w.r.t. Baseline in \cite{zafar2017fairness}
& & & & & & x & & & & & & & & & x & & & & & & & & &\\
\hline
Exp. w.r.t. \cite{kamiran2009classifying}
& & & & & & & & & x & & & & & & & & & & & x & & x & &\\
\hline
Exp. w.r.t. \cite{kamishima2011fairness}
& & & & & & & & & x & & & & & & & & x & & & x & & & &\\
\hline
Exp. w.r.t. \cite{kamiran2010classification}
& & & & & & & & & & & & & & & & & x & & & & & & &\\
\hline
Exp. w.r.t. \cite{zemel2013learning} 
& & & & & & & & & & & & & & & & & & & & x & & & &\\
\hline
Code Available
& x & & & & & x & & x & & & & x & & & x & & x & & & & x & & &\\
\hline
\hline
\end{tabular}
\caption{A summary of the characteristics of the different methods concerning fairness. The symbol 'x' means the presence of a property (row) for a specific method (column). x$^{*}$: the theoretical results however do not correspond to their convex method.}
\label{tab:review}
\end{table}

\section{Datasets}
\label{app:dataset}
In the following the datasets used in Section~\ref{sec:exps} are presented, outlining their tasks, type of features and source of data. Table~\ref{tab:datasets} provide a summary of the datasets statistics.

\begin{itemize}
	\item \emph{Arrhythmia}: from UCI repository, this database contains 279 attributes concerning the study of H. Altay Guvenir. The aim is to distinguish between the presence and absence of cardiac arrhythmia and to classify it in one of the 16 groups. In our case, we changed the task with the binary classification between "Class 01" (i.e. "Normal") against the other 15 classes (different classes of arrhythmia).
 	\item \emph{COMPAS} (Correctional Offender Management Profiling for Alternative Sanctions): it is a popular commercial algorithm used by judges and parole officers for scoring criminal defendant’s likelihood of reoffending (recidivism). It has been shown that the algorithm is biased in favor of white defendants based on a 2 year follow up study. This dataset contains variables used by the COMPAS algorithm in scoring defendants, along with their outcomes within 2 years of the decision, for over 10000 criminal defendants in Broward County, Florida. In the original data, 3 subsets are provided. We concentrate on the one that includes only violent recividism\footnote{Analysis of the recidivism COMPAS dataset: www.propublica.org/article/how-we-analyzed-the-compas-recidivism-algorithm}.
 \item \emph{Adult}: from UCI repository, this database contains 14 features concerning demographic characteristics of $45222$ instances ($32561$ for training and $12661$ for test). The task is to predict if a person has an income per year that is more (or less) than $50000\,\$$. Concerning the Adult dataset we used the provided training and test sets.
 \item \emph{German}: it is a dataset where the task is to classify people described by a set of 20 features (7 numerical, 13 categorical) as good or bad credit risks. The features are related to the economical situation of the person, as for example: credit history and amount, saving account and bonds, year of the present employment, property and others. Moreover, a set of features is concerning personal information, e.g. age, gender, if the person is a foreign, and personal status.
 \item \emph{Drug}: this dataset contains records for 1885 respondents. Each respondent is described by 12 features: Personality measurements which include NEO-FFI-R (neuroticism, extraversion, openness to experience, agreeableness, and conscientiousness), BIS-11 (impulsivity), and ImpSS (sensation seeking), level of education, age, gender, country of residence and ethnicity. All input attributes are originally categorical and are quantified. After quantification values of all input features can be considered as real-valued. In addition, participants were questioned concerning their use of 17 legal and illegal drugs and one fictitious drug (Semeron) which was introduced to identify over-claimers. For each drug, the respondents have to select one of the answers: never used the drug, used it over a decade ago, or in the last decade, year, month, week, or day. In this sense, this dataset contains 18 classification problems, each one with seven classes: "Never Used", "Used over a Decade Ago", "Used in Last Decade", "Used in Last Year", "Used in Last Month", "Used in Last Week", and "Used in Last Day". We make the problem number $16$ (concerning heroin) a binary problem by exploiting the task "Never used" versus "Others" (i.e. "Used").
\end{itemize} 

\begin{table*}[t!]
\small
\centering
\begin{tabular}{|l|c|c|c|}
\hline
\hline
Dataset		&	Examples	&	Features	&	Sensitive Variable	\\
\hline
\hline
Arrhythmia	&	452			 &	279			&	Gender		\\
COMPAS		&	6172		 &	10			&	Ethnicity	\\
Adult		&	32561, 12661 &	12			&	Gender		\\
German 		& 	1700		 &	20			& 	Foreign 	\\
Drug		&	1885 		 &	11			& 	Ethnicity 	\\
\hline
\hline
\end{tabular}
\caption{Datasets statistics (for Adult we reported the number of train testing examples provided) and their sensitive feature. Gender considers the two groups as male and female; ethnicity considers the ethnic groups white and other ethnic groups; foreign considers being or not being a foreign person.}
\label{tab:datasets}
\end{table*}

\section{Varying the Value of $\epsilon$}
\label{app:epsilon}
In this section we present a set of experiments, as a proof of concept, that our selection of $\epsilon = 0$ for our method is reasonable and study the impact of different values of $\epsilon$ have concerning DEO and accuracy performance.

\begin{figure}
\centering
\includegraphics[width=0.50\columnwidth]{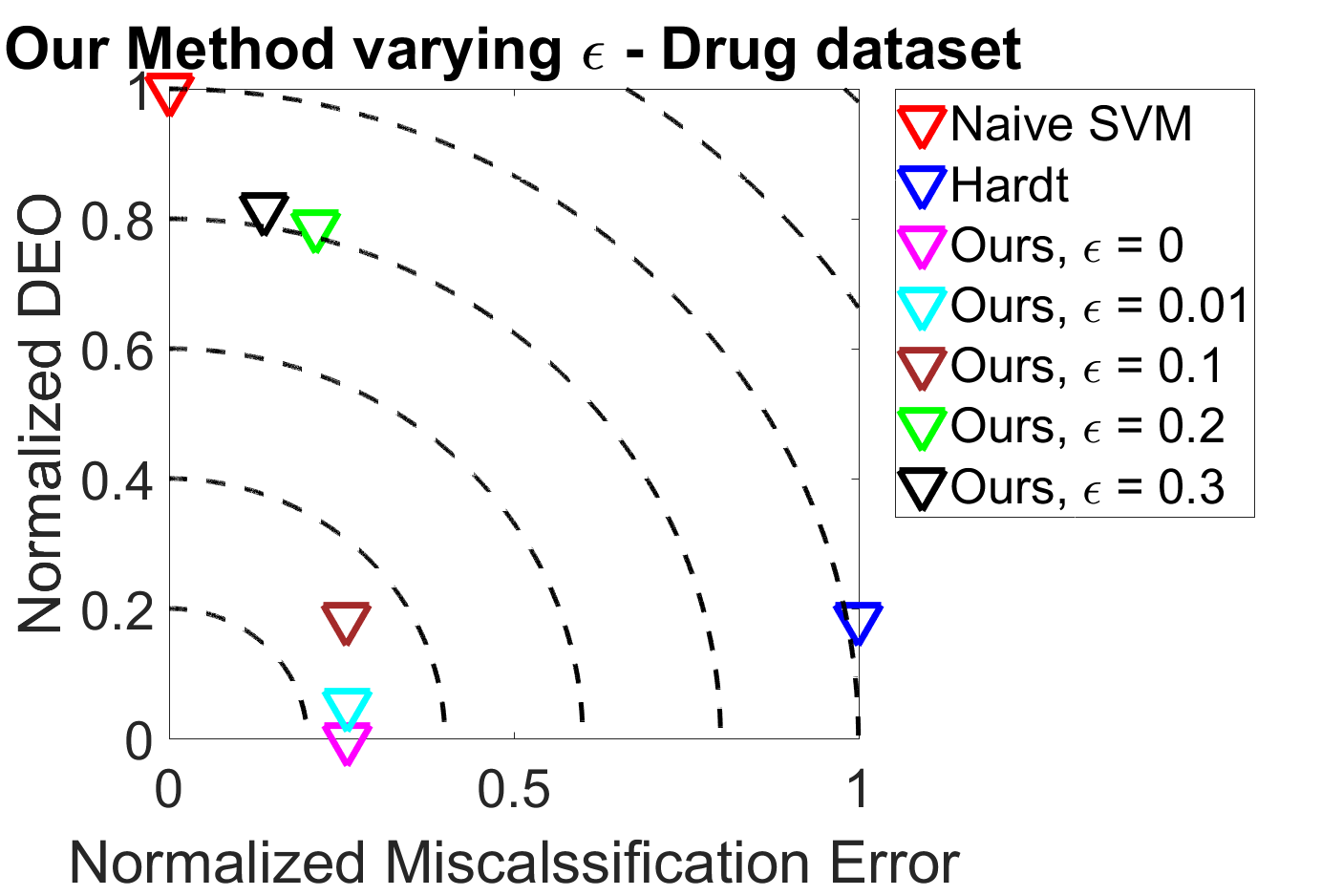}
\caption{Results concerning the Drug dataset for Na\"{i}ve SVM, Hard method and our method with different values of $\epsilon$.}
\label{fig:drugeps}
\end{figure}

We follow the same experimental setting presented in Section~\ref{sec:exps} for the Drug dataset, implementing our nonlinear method with $\epsilon$ equals to $0, 0.01, 0.1, 0.2, 0.3$. The results of this experiment are presented in Figure~\ref{fig:drugeps}, where we show also the results for Na\"{i}ve SVM and Hard method. It is possible to note how increasing the value of $\epsilon$, our model has smaller error but stronger unfairness (i.e. higher DEO). 

{\section{Visualization of the results of Table~\ref{tab:results}}
In Figure~\ref{fig:tableexplanation2} we reported the equivalent of Figure~\ref{fig:tableexplanation} for the case when $s$ is not inside $\boldsymbol{x}$.
Note that we can reach the same conclusions drown for Table~\ref{tab:results} and Figure~\ref{fig:tableexplanation}.
\begin{figure}
\centering
\includegraphics[width=0.45\columnwidth]{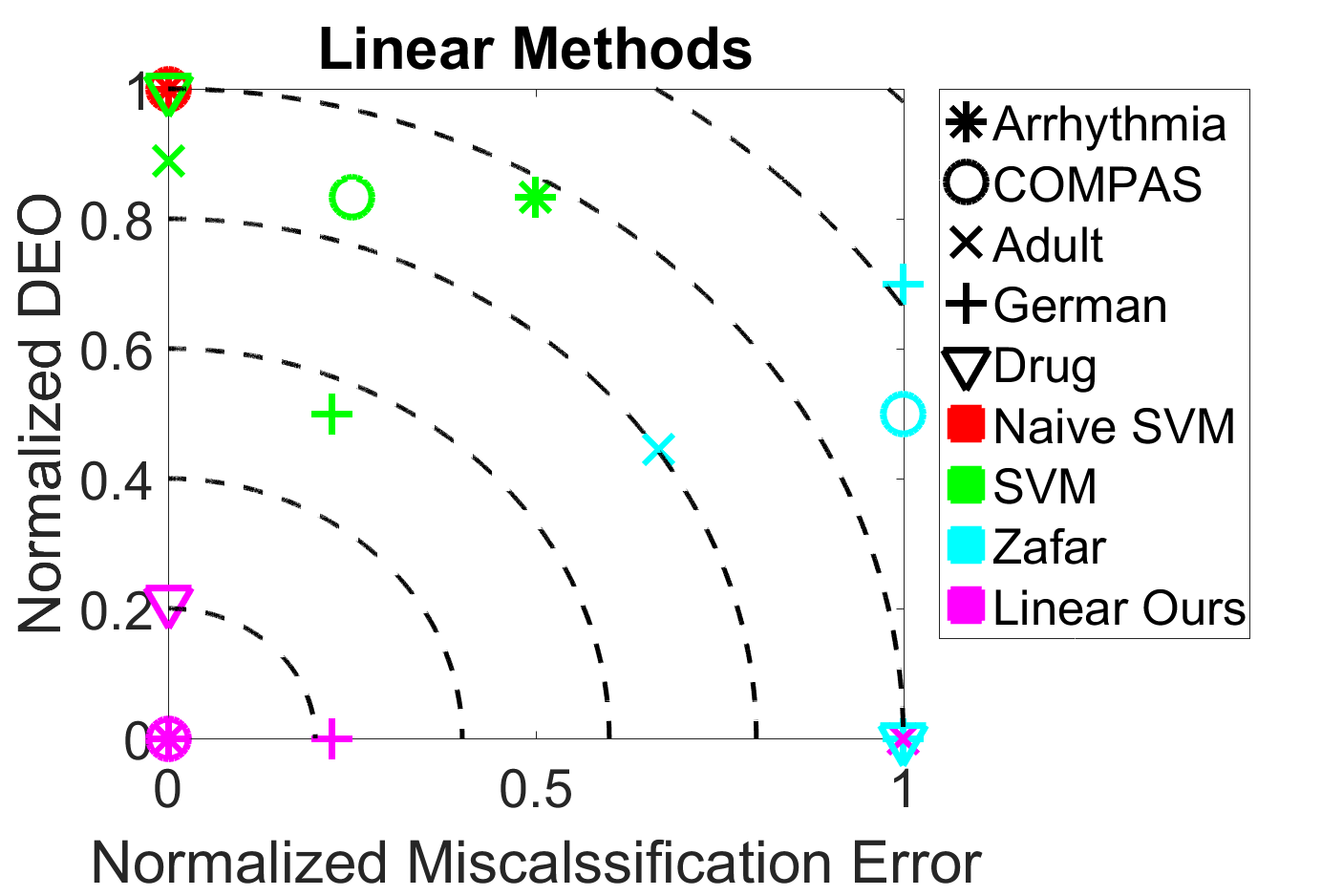} \quad
\includegraphics[width=0.45\columnwidth]{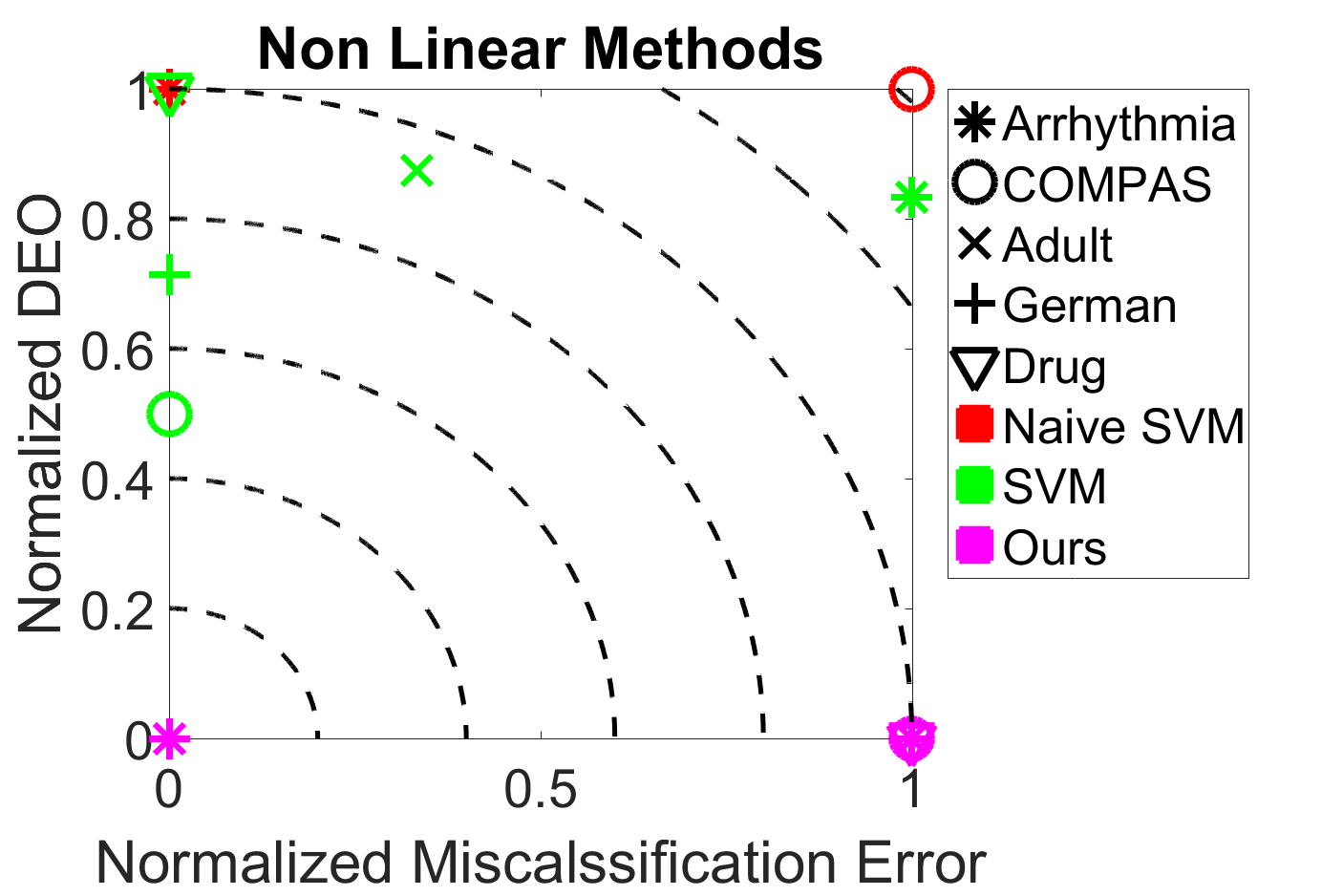}
\caption{{Results of Table~\ref{tab:results} of linear (left) and nonlinear (right) methods, when the error and the DEO are normalized in $[0,1]$ column-wise and when $s$ is not inside $\boldsymbol{x}$. Different symbols and colors refer to different datasets and method respectively. The closer a point is to the origin, the better the result is.}}
\vspace{-.4cm}
\label{fig:tableexplanation2}
\end{figure}}

\section{Approximation of the DEO}
\label{app:exp_approximation_DEO}
\begin{figure}
\centering
\includegraphics[trim={1.5cm 1.8cm 1.8cm 2cm},clip,width=0.3\textwidth]{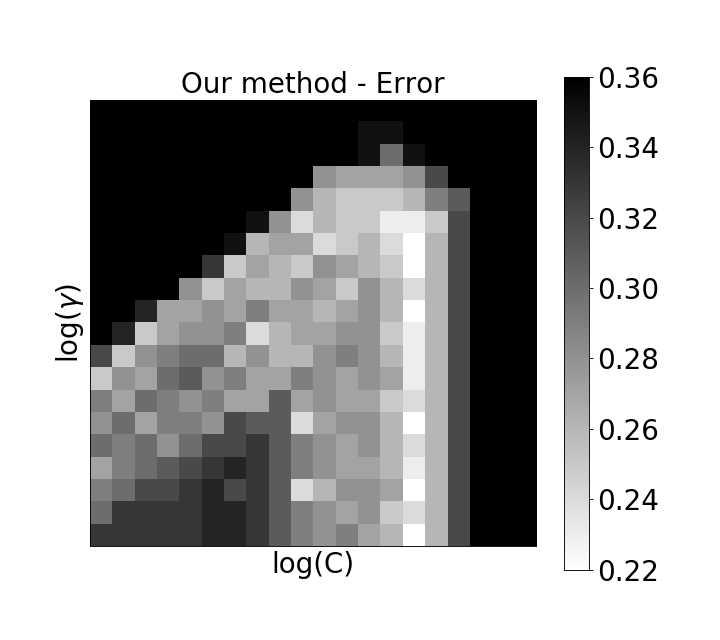}
\includegraphics[trim={1.5cm 1.8cm 2.5cm 2cm},clip,width=0.3\textwidth]{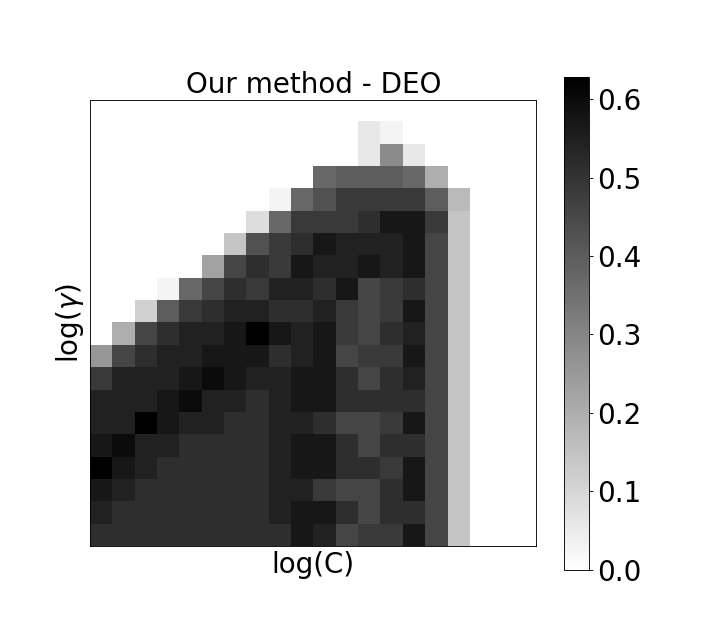}
\includegraphics[trim={1.5cm 1.8cm 2.5cm 2cm},clip,width=0.3\textwidth]{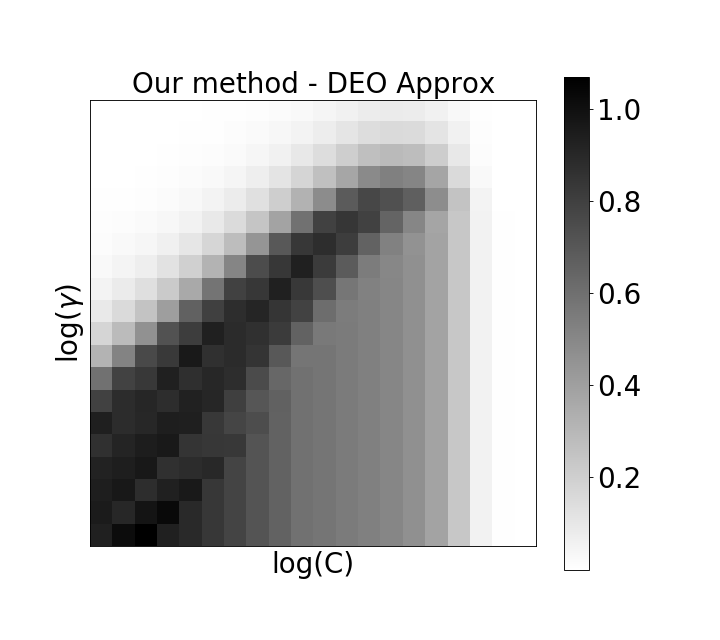}
\caption{Error (Left), DEO (Middle) and our approximation of the DEO (Right) for the German dataset by using our method. For the three images, the x-axis is the logarithm of the regularization hyperparameter $C$ and the y-axis is the logarithm of the RBF hyperparameter $\gamma$.}
\label{fig:approxdeo}
\end{figure}

In this section, we numerically show the difference between the DEO and our approximation of it. Figure~\ref{fig:approxdeo} compares the DEO with our approximation of the DEO and the classification error. We collected these results for the German dataset on the validation set, changing the two hyperparameters $C$ and $\gamma$ (in the nonlinear case).
We can note how our approximation of the DEO is empirically similar to the original DEO. It is interesting to highlight that, a correct approximation of the DEO is particularly important where the error is low. 

\section{Dual Problem for SVM with Fairness Constraint}
\label{app:dual}
We follow the usual approach to derive the dual problem for SVMs, which uses the method of Lagrange multipliers~\cite{vapnik1998statistical}. We define the 
Lagrangian function 
\begin{eqnarray}
\nonumber
{\mathcal L}(\boldsymbol{w},\boldsymbol{\xi},\boldsymbol{\alpha},\boldsymbol{\beta},\boldsymbol{\rho}) {=} \frac{1}{2} \langle \boldsymbol{w},\boldsymbol{w} \rangle + C \sum_{i=1}^n \xi_i - \sum_{i=1}^n \alpha_i (y_i \langle \boldsymbol{\phi}(\boldsymbol{x}_i), \boldsymbol{w} \rangle - 1 + \xi_i) - \beta_i \xi_i +~~~~~~~~ \\ 
~~~~~~~~~~~\rho_1(\langle \boldsymbol{w},\boldsymbol{u} \rangle - \epsilon) -\rho_2(\langle \boldsymbol{w},\boldsymbol{u} \rangle + \epsilon)
\label{eq:Lag}
\end{eqnarray}
where $\boldsymbol{\alpha},\boldsymbol{\beta}$ and $\boldsymbol{\rho}$ are the vector of Lagrange multipliers and are constrained to be nonnegative. We set the derivative of the Lagrangian with respect to the primal variables $\boldsymbol{w}$ and $\boldsymbol{\xi}$ equal to zero. In the latter case we obtain that 
\begin{equation}
C - \alpha_i - \beta_i = 0
\label{eq:d11}
\end{equation}
from which we can remove the variable $\beta_i$ in place of the constraint $\alpha_i \leq C$. In the former case we obtain the expression for $\boldsymbol{w}$,
\begin{equation}
\boldsymbol{w} = \sum_{i=1}^n \alpha_i y_i \boldsymbol{\phi}(\boldsymbol{x}_i) + (\rho_1-\rho_2) \boldsymbol{u}.
\label{eq:d22}
\end{equation}
Using~\eqref{eq:d11} and~\eqref{eq:d22} in~\eqref{eq:Lag} we obtain the expression
\begin{equation}
- \frac{1}{2} \Big\|\sum_{i=1}^n \alpha_i y_i \boldsymbol{\phi}(\boldsymbol{x}_i) + (\rho_1-\rho_2) \boldsymbol{u} \Big\|^2 + \sum_{i=1}^n \alpha_i - \epsilon (\rho_1+ \rho_2).
\label{eq:Dual}
\end{equation}
The dual problem is then to maximize this quantity subject to the constraints that $\alpha_i \in [0,C]$ and $\rho_1,\rho_2 \geq 0$.

The KKT conditions are
\begin{eqnarray}
\alpha_i ( 1 - y_i \langle \boldsymbol{w},\boldsymbol{\phi}(\boldsymbol{x}_i) \rangle - 1 + \xi_i) & = & 0 \\
C-\alpha_i \xi_i & = & 0 \\
\rho_1(\langle \boldsymbol{w}, \boldsymbol{u} \rangle - \epsilon) & = & 0 \\
\rho_2(\langle \boldsymbol{w}, \boldsymbol{u} \rangle + \epsilon) & = & 0.
\end{eqnarray}
Clearly at most one of the variables $\rho_1$ and $\rho_2$ can be strictly positive. We may then let $\rho = \rho_1-\rho_2$ and rewrite the objective function as 
\begin{equation}
- \frac{1}{2} \Big\|\sum_{i=1}^n \alpha_i y_i \boldsymbol{\phi}(\boldsymbol{x}_i) + \rho \boldsymbol{u} \Big\|^2 + \sum_{i=1}^n \alpha_i - \epsilon |\rho|
\label{eq:Dual2}
\end{equation}
and optimize over $\boldsymbol{\alpha} \in [0,C]^n$ and $\rho\in \mathbb{R}$. It is interesting to study this problem when $\epsilon = 0$. In this case we can easily solve for $\rho$ obtaining the simplified objective
$$
- \frac{1}{2} \Big\| \sum_{i=1}^n \alpha_i y_i (I-P) \boldsymbol{\phi}(\boldsymbol{x}_i) ) \Big\|^2 + \sum_{i=1}^n \alpha_i
$$
where $P$ is the orthogonal projection along the direction of $\boldsymbol{u}$, that is $P= \frac{\boldsymbol{u}}{\| \boldsymbol{u} \|} \otimes \frac{\boldsymbol{u}}{\| \boldsymbol{u} \|}$.
This is equivalent to use the standard SVM with the kernel 
\[
{\widetilde \kappa}(\boldsymbol{x},\boldsymbol{t}) = \langle \boldsymbol{\phi}(\boldsymbol{x}), (I-P) \boldsymbol{\phi}(\boldsymbol{t})\rangle = \kappa(\boldsymbol{x},\boldsymbol{t}) -\frac{\langle \boldsymbol{x}, \boldsymbol{u} \rangle \langle \boldsymbol{t}, \boldsymbol{u} \rangle}{\| \boldsymbol{u} \|^2}
\]
In particular if $\displaystyle \boldsymbol{u} = \frac{1}{n_a} \sum_{i\in \mathcal{I}^{+,a}} \boldsymbol{x}_i -\frac{1}{n_b} \sum_{i\in \mathcal{I}^{+,b}} \boldsymbol{x}_i$, we obtain 
$$
{\widetilde \kappa}(\boldsymbol{x},\boldsymbol{t}) =\kappa(\boldsymbol{x},\boldsymbol{t}) -\frac{\frac{1}{n_a}\sum\limits_{i\in \mathcal{I}^{+,a}} \kappa(\boldsymbol{x},\boldsymbol{x}_i) -\frac{1}{n_b} \sum\limits_{i\in \mathcal{I}^{+,b}} \kappa(\boldsymbol{x},\boldsymbol{x}_i)}{\frac{1}{n_a^2}\sum\limits_{i,j \in \mathcal{I}^{+,a}} \kappa(\boldsymbol{x}_i,\boldsymbol{x}_j) + \frac{1}{n_b^2}\sum\limits_{i,j \in \mathcal{I}^{+,b}} \kappa(\boldsymbol{x}_i,\boldsymbol{x}_j) 
- \frac{2}{n_an_b}\sum\limits_{i \in \mathcal{I}^{+,a}}\sum\limits_{j \in \mathcal{I}^{+,b}} \kappa(\boldsymbol{x}_i,\boldsymbol{x}_j)}.
$$
This new kernel can then be interpreted as a change of feature mapping $\boldsymbol{x} \mapsto (I-P) \boldsymbol{\phi}(\boldsymbol{x}) = \boldsymbol{\phi}(\boldsymbol{x}) - \langle \boldsymbol{\phi}(\boldsymbol{x}), \frac{\boldsymbol{u}}{\| \boldsymbol{u} \|} \rangle \frac{\boldsymbol{u}}{\| \boldsymbol{u} \|}$.

As a final remark, we note that for other proper convex loss functions (e.g. square loss or logistic loss) the dual problem can be derived via Fenchel duality~\citep[see e.g.][]{Rockafellar1970}. We leave the full details to a future occasion.

\section{Multiple Valued Sensitive Features}
\label{sec:luca:th:FK:MMF}
Our method presented in Section~\ref{sec:luca:th:FK} can be naturally extended to the case that the sensitive variable takes multiple categorical values, that is $s \in \{g_1,\dots,g_k\}$ for some $k \geq 2$.
In particular, when $\epsilon = 0$, the fairness constraint in Problem~\eqref{eq:alg:empirical} requires that 
\begin{align}
\hat{L}^{+,g_1}(f) = \hat{L}^{+,g_2}(f) = \cdots = \hat{L}^{+,g_k}(f).
\end{align}
Furthermore if the linear loss function is used, these constraints becomes%
\[
\langle \mathbf{w} , \mathbf{u}_{1} - \mathbf{u}_g \rangle = 0 , \quad \forall g \in \{g_2, \cdots, g_k\}
\]
where we defined, for $g\in \{g_1,\dots,g_k\}$
\[
\mathbf{u}_g= \frac{1}{n^{+,g}} \sum_{ i \in \mathcal{I}^{+,g}}
\mathbf{\phi}(\boldsymbol{x}_i),
\]
with $\mathcal{I}^{+,g} = \{i: y_i = 1, s = g \}$ and $n^{+,g} = |\mathcal{I}^{+,g}|$.
Thus, we need to satisfy $k-1$ orthogonality constraints which try to enforce a balance between the different sensitive groups as measured by the barycenters of the within groups positive labeled points. Similar considerations apply when dealing with multiple sensitive features.

\begin{table}[t!] 
\small
\centering
\begin{tabular}{|l|c|c|c|}
\multicolumn{4}{c}{Arrhythmia dataset} \\
\hline
\hline
Method & Accuracy & DEO & Selected Features \\
\hline
\hline
Na\"{i}ve Lin. SVM & $0.79 \pm 0.06$ & $0.14 \pm 0.03$ & - \\
Linear SVM & $0.78 \pm 0.07$ & $0.13 \pm 0.04$ & - \\
\hline
Na\"{i}ve Lasso & $0.79 \pm 0.07$ & $0.11 \pm 0.04$ & $ 22.7 \pm 9.1$ \\
Lasso 			& $0.74 \pm 0.04$ & $0.07 \pm 0.04$ & $ 5.2 \pm 3.7$ \\
Hardt 			& $0.71 \pm 0.05$ & $0.04 \pm 0.06$ & $ 5.2 \pm 3.7$ \\
Our Lasso 		& $0.77 \pm 0.02$ & $0.03 \pm 0.02$ & $ 7.5 \pm 2.0$ \\
\hline
\hline
\multicolumn{4}{c}{}\\
\multicolumn{4}{c}{Arrhythmia dataset - Training set reduced by 50\%} \\
\hline
\hline
Method & Accuracy & DEO & Selected Features \\
\hline 
\hline
Na\"{i}ve Lin. SVM & $0.69 \pm 0.03$ & $0.16 \pm 0.03$ & - \\
Linear SVM & $0.68 \pm 0.03$ & $0.15 \pm 0.03$ & - \\
\hline
Na\"{i}ve Lasso & $0.73 \pm 0.04$ & $0.15 \pm 0.06$ & $ 14.1 \pm 6.6$ \\
Lasso 			& $0.70 \pm 0.04$ & $0.09 \pm 0.05$ & $ 7.9 \pm 8.0$ \\
Hardt 			& $0.67 \pm 0.06$ & $0.08 \pm 0.07$ & $ 7.9 \pm 8.0$ \\
Our Lasso 		& $0.71 \pm 0.04$ & $0.03 \pm 0.04$ & $ 9.0 \pm 7.3$ \\
\hline
\hline
\end{tabular}
\vspace{.1truecm}
\caption{Results (average $\pm$ standard deviation) when the model is the Lasso, concerning accuracy, DEO and the number of features with weight bigger than $10^{-8}$ over the $279$ original features. The experiment has been repeated also reducing the training set.}
\label{tab:results_lasso} 
\end{table}

\begin{table}[th]
\small
\centering
\begin{tabular}{|l|c|}
\hline
\hline
Dataset			& $\hat{\Delta}$ \\
\hline
\hline
Toytest			& 0.03 \\
Toytest Lasso 	& 0.02 \\
\hline
Arrhythmia		& 0.03 \\
COMPAS			& 0.04 \\
Adult			& 0.06 \\
German 			& 0.05 \\
Drug			& 0.03 \\
\hline
\hline
\end{tabular}
\caption{The $\hat{\Delta}$ for the exploited datasets. A smaller $\hat{\Delta}$ means a better approximation of the DEO in our method.}
\label{tab:delta}
\end{table}

\end{document}